\documentclass[10pt,twocolumn,letterpaper]{article}

\usepackage[pagenumbers]{cvpr} 
\usepackage[]{cvpr}    
\usepackage{times}
\usepackage{epsfig}
\usepackage{graphicx}
\usepackage{amsmath}
\usepackage{amssymb}
\usepackage{xcolor}         
\usepackage{float}
\usepackage{caption}
\usepackage{amsfonts,amssymb}
\usepackage{mathrsfs}
\usepackage{tabularx}
\usepackage{wrapfig}
\usepackage{verbatim}
\usepackage{dsfont}
\usepackage{tablefootnote}
\usepackage[stable]{footmisc}
\usepackage{booktabs}
\usepackage{multicol}
\usepackage{multirow}
\usepackage{color}
\usepackage{subcaption}
\usepackage{pythonhighlight}
\usepackage{algpseudocode}
\usepackage{algorithm}
\usepackage{physics}
\usepackage{url}

\definecolor{todocolor}{RGB}{255,0,00}

\definecolor{jiapeng}{rgb}{0.2, 0.4,0.9}

\renewcommand{\vec}[1]{\mathbf{#1}}
\newcommand{\set}[1]{\mathcal{#1}}
\newcommand{\expnumber}[2]{{#1}\mathrm{e}{#2}}
\renewcommand{\paragraph}[1]{\smallskip\noindent\textbf{#1}}

\definecolor{sim}{RGB}{128,0,128}
\definecolor{interp}{RGB}{255, 169, 0}
\definecolor{mytbcol}{RGB}{175,227,246}

\DeclareMathOperator{\rec}{rec}
\DeclareMathOperator{\view}{view}

\DeclareMathOperator{\img}{img}

\DeclareMathOperator{\pos}{pos}
\DeclareMathOperator{\scale}{scale}

\definecolor{cvprblue}{rgb}{0.21,0.49,0.74}
\usepackage[pagebackref,breaklinks,colorlinks,allcolors=cvprblue]{hyperref}

\definecolor{tabfirst}{rgb}{1, 0.7, 0.7} 
\definecolor{tabsecond}{rgb}{1, 0.85, 0.7} 
\definecolor{tabthird}{rgb}{1, 1, 0.7} 
\newcommand{\tabfirst}[1]{\colorbox{tabfirst}  {#1}}
\newcommand{\tabsecond}[1]{\colorbox{tabsecond} {#1}}

\def\paperID{4288} 
\def\confName{CVPR}
\def\confYear{2025}

\title{GAF:  Gaussian Avatar Reconstruction from Monocular Videos \\ via Multi-view Diffusion \vspace{-0.2cm} } 
\author{ Jiapeng Tang$^1$ \quad Davide Davoli$^2$ \quad Tobias Kirschstein$^1$ \quad Liam Schoneveld$^3$ \quad Matthias Nie{\ss}ner$^1$ \vspace{0.15cm} \\
     {
        \renewcommand{\arraystretch}{0.2} 
        \begin{tabular}[t]{c c c}%
          $^{1}$Technical University of Munich &
            $^{2}$Toyota Motor Europe NV/SA &
            $^{3}$ Woven by Toyota \\
            &  \footnotesize{ associated partner by contracted service} &
        \end{tabular}
     }  
}

\begin{document}
\twocolumn[{%
\renewcommand\twocolumn[1][]{#1}%
\maketitle 
\begin{center}
    \vspace{-0.88cm} 
    \centering
    \captionsetup{type=figure}
    \includegraphics[width=\linewidth]{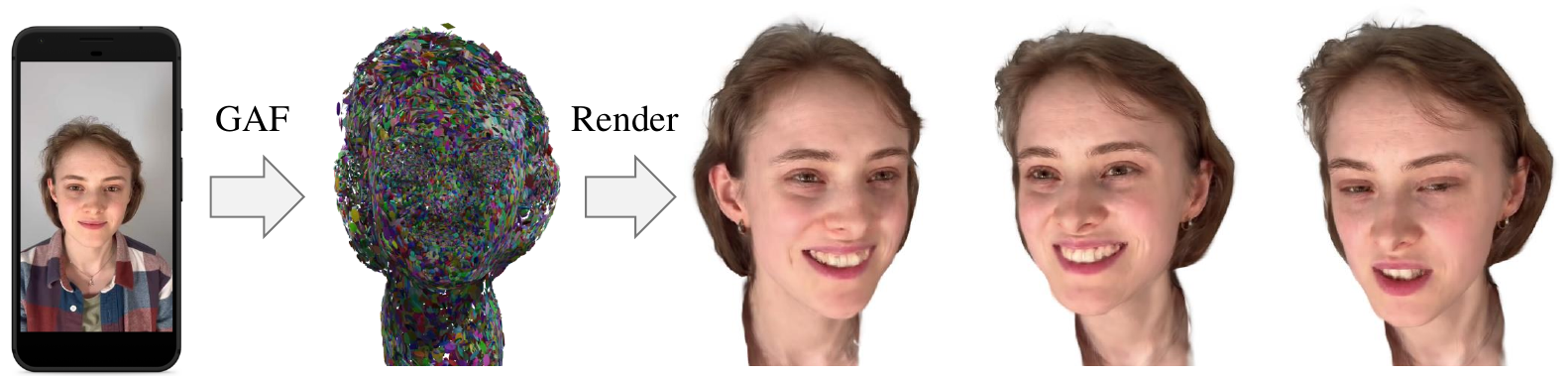}
    \vspace{-6mm}
  \caption{Given a short, monocular video captured by a commodity device such as a smartphone, GAF reconstructs a 3D Gaussian head avatar, which can be re-animated and rendered into photo-realistic novel views. Our key idea is to distill the reconstruction constraints from a multi-view head diffusion model  in order to extrapolate to unobserved views and expressions.  
  }
    \label{fig:teaser}
\end{center}
}]
\begin{abstract}
We propose a novel approach for reconstructing animatable 3D Gaussian avatars from monocular videos captured by commodity devices like smartphones. Photorealistic 3D head avatar reconstruction from such recordings is challenging due to limited observations, which leaves unobserved regions under-constrained and can lead to artifacts in novel views.
To address this problem, we introduce a multi-view head diffusion model, leveraging its priors to fill in missing regions and ensure view consistency in Gaussian splatting renderings. To enable precise viewpoint control, we use normal maps rendered from FLAME-based head reconstruction, which provides pixel-aligned inductive biases. We also condition the diffusion model on VAE features extracted from the input image to preserve facial identity and appearance details.
For Gaussian avatar reconstruction, we distill multi-view diffusion priors by using iteratively denoised images as pseudo-ground truths, effectively mitigating over-saturation issues.  To further improve photorealism, we apply latent upsampling priors to refine the denoised latent before decoding it into an image.
We evaluate our method on the NeRSemble dataset, showing that
GAF outperforms previous state-of-the-art methods in novel view synthesis. Furthermore, we demonstrate higher-fidelity avatar reconstructions from monocular videos captured on commodity devices.  \href{https://tangjiapeng.github.io/projects/GAF}{Project Page: https://tangjiapeng.github.io/projects/GAF}
\end{abstract}


\section{Introduction}
Creating photorealistic and animatable head avatars has long been a challenge in computer vision and graphics.
This is crucial for a vast variety of applications such as immersive telepresence in virtual and augmented reality, computer games, movie effects, virtual classrooms, and videoconferencing. 
Here, the goal is to generate avatars that can be realistically rendered from any viewpoint, accurately capturing facial geometry and appearance, while also enabling easy animation for realistic head portrait videos depicting various expressions and poses.
The democratization of high-fidelity head avatars from commodity devices is a challenge of widespread interest. However, photo-realistic head avatar reconstruction from monocular videos is challenging and ill-posed, due to limited head observations causing a lack of constraints for novel-view rendering.
%

To create photorealistic, animatable human avatars, researchers have integrated NeRF~\cite{nerf,instantngp} and Gaussian Splatting (GS)~\cite{GaussianSplatting} techniques with parametric head models~\cite{FLAME, NPHM, DPHM}, enhancing control over unseen poses and expressions. Approaches like~\cite{AvatarMAV, Gaussianheadavatar, GaussianAvatars, NPGA} achieve high-quality head reconstructions and realistic animations using dense multi-view datasets, typically captured in controlled studio settings. However, these methods encounter significant limitations with monocular recordings from commodity cameras, such as smartphone portrait videos. 
Monocular methods~\cite{Nerface, IMAvatar, MonoAvatar, PointAvatar, Monogaussianavatar, Splattingavatar, INSTA, FlashAvatar} aim to reconstruct head avatars from single-camera videos but often rely on substantial head rotations to capture various angles. They only reconstruct visible regions from input frames, leaving occluded areas incomplete and under-constrained.  For instance, as illustrated in Fig.~\ref{fig:novelview}
, front-facing videos provide limited capture on side regions, leading to visible artifacts when rendering extreme side views.

To this end, we introduce a multi-view head diffusion model that learns the joint distribution of multi-view head images. 
Given a single-view input image, our model generates a set of view-consistent output images. By leveraging view-consistent diffusion priors on human heads, our approach enables the robust reconstruction of Gaussian avatars, preserving both appearance and identity across novel perspectives. 
Unlike existing methods~\cite{3DiM, Zero123, MvDream, ImageDream} that use camera pose as a conditioning factor for viewpoint control, we use the normal map rendered from a reconstructed parametric face model as guidance. The normal maps provide stronger inductive biases, offering more precise and reliable novel view generation for heads.
Specifically, we condition our multi-view diffusion process on 2D features extracted from the input image’s autoencoder, rather than only using high-level semantic vectors like CLIP embeddings~\cite{CLIP}. This design allows us to incorporate fine-grained identity and appearance details directly into the multi-view denoising process, ensuring that the generated images maintain coherence and consistency across viewpoints in terms of identity and appearance.

To exploit multi-view diffusion priors for Gaussian head reconstruction, we employ a diffusion loss that uses iteratively denoised images as pseudo-ground truths for novel view renderings, instead of using a single-step score distillation sampling loss~\cite{Dreamfusion}.
Compared to NeRF, 3D-GS enables real-time multi-view rendering, significantly reducing optimization time.
Moreover, to improve the fidelity of the Gaussian renderings, we introduce a latent upsampler model to enhance facial details within the denoised latent before decoding it back to image space. As demonstrated in Fig.~\ref{fig:teaser}, our approach generates high-quality, photorealistic novel views of head avatars using only short monocular videos captured by a smartphone. Extensive comparisons with state-of-the-art methods show that our approach delivers higher fidelity and more view-consistent avatar reconstructions from monocular videos. Our GAF significantly outperforms state-of-the-art methods in novel view and expression synthesis on the NeRSemble dataset and other monocular datasets captured by commodity devices.
Our contributions can be summarized as follows:
\begin{itemize}
    \item 
    We introduce a novel approach for reconstructing photorealistic, animatable head avatars from monocular videos with limited coverage captured on commodity devices, using multi-view head diffusion priors to regularize and complete unobserved regions.
    \item We propose a multi-view head diffusion model that generates consistent multi-view images from a single-view input, guided by normal maps rendered from FLAME head reconstructions to improve viewpoint control.
    \item We present {a strategy to enhance} the photorealism and cross-view consistency of Gaussian splatting by integrating latent upsampling and multi-view diffusion priors. 
\end{itemize}

\section{Related Work}
\paragraph{3D Scene Representations.} Neural radiance fields ~\cite{nerf} and its variants~\cite{instantngp, plenoxels, tensorRF, Mip-nerf, MipNerf360, ZipNerf, Mesh2NeRF} revolutionized 3D scene reconstruction from multi-view images.
However, NeRF-based methods are often hindered by computational inefficiency during both training and inference stages. Gaussian Splatting~\cite{GaussianSplatting} represents a scene as a composition of discrete geometric primitives,~\ie 3D Gaussians, and employs an explicit rasterizer for rendering. Compared to NeRF, GS has achieved notable runtime speedups in  the training and inference stages. This enables real-time performance and more favorable image synthesis. Unlike polygonal meshes, which require careful topology handling, GS supports substantial topology changes, making it more adaptable to complex surfaces and varying densities.
%

\paragraph{Parametric Face/Head Models.}
Based on statistical priors of 3D morphable models (3DMM)\cite{blanz2023morphable, vlasic2006face, FLAME, guo20213d, tran2018nonlinear}, many works~\cite{face2face, NeuralHeadAvatars, ROME, DECA} learn 3D face/head reconstruction and tracking from single RGB images/videos. More recent methods~\cite{NPHM, ImFace, IMAvatar, DPHM} leveraged signed distance fields~\cite{DeepSDF} and deformation fields~\cite{Oflow, LPDC, NSDP} based on coordinate-MLPs for more fine-grained geometry reconstruction including hair and beards. 
HeadGAP~\cite{HeadGAP} and GPHM~\cite{GPHM} learned parametric models for head Gaussians.
Our work uses the VHAP tracker~\cite{qian2024versatile} to obtain coarse geometries as guidance for dynamic avatar reconstruction.

\paragraph{Photo-realistic Avatar Reconstruction.}
%
%
To create photo-realistic animatable head avatars, several works~\cite{Nerface, INSTA, H3QAvatar, INSTA, PointAvatar, FlashAvatar, Splattingavatar, GaussianAvatars, Gaussianheadavatar} have combined NeRF/GS with 3D morphable models (3DMM). %
%
Our work is closely related to animatable Gaussian splats~\cite{GaussianAvatars, Splattingavatar}, which attached splats to the triangles of FLAME mesh, and updated their properties by triangle deformations controlled by FLAME parameters.
Although promising results have been achieved, they typically require multi-view setups with high-quality cameras in studio environments~\cite{MVP, Nersemble}. Some works reconstruct avatars from monocular videos~\cite{PointAvatar, MonoAvatar, INSTA, FlashAvatar}. However, they only reconstruct the visible region from inputs, lacking effective priors to complete missing areas. 
P4D~\cite{Portrait4D}, P4D-v2~\cite{Portrait4Dv2}, GAGAvatar~\cite{GAGAvatar}, and MGGTalk~\cite{MGGTalk} use feed-forward networks to predict expression-driven 3D heads represented by Gaussians or triplane features~\cite{peng2020convolutional}. While achieving impressive head reenactments, they could exhibit obvious artifacts in novel views. To address this, we utilize multi-view diffusion priors to complete unobserved regions of the face, ensuring photorealistic renderings from extremely novel viewpoints. 

\paragraph{Distillation 2D Priors for 3D Generation.}
Diffusion models~\cite{song2019generative, song2020improved, song2020score, ho2020denoising} have been widely applied to 3D asset generation~\cite{zhang20233dshape2vecset,tang2024diffuscene,cao2024motion2vecsets, vahdat2022lion,luo2021diffusion,zhou20213d}. Some methods leverage large-scale pretrained text-to-image priors~\cite{LatentDiffusionModels, Imagen, DALLE2} for 3D synthesis~\cite{Magic3D, Dreamgaussian} using score distillation sampling (SDS) loss~\cite{Dreamfusion} and its variants~\cite{ProlificDreamer, CSDLoss, Realfusion}.
Several studies~\cite{HeadSculpt, DreamAvatar, HeadStudio} applied SDS loss for text-driven 3D head avatar generation. For single-view 3D reconstruction, approaches like RelFusion~\cite{Realfusion}, Magic123~\cite{Magic123}, and Dreambooth3D~\cite{Dreambooth3d} adapt text-to-image diffusion priors to specific objects to preserve identity. However, achieving consistent novel views from text prompts alone remains challenging without leveraging the input image.
To address this, we learn multi-view image diffusion priors conditioned on a single image. Rather than using a single-step SDS loss, we use iteratively denoised images as pseudo-ground truths for novel view supervision, similar to ReconFusion~\cite{ReconFusion}. 
We also introduce pretrained latent upsampler diffusion priors to improve photo-realism of pseudo-ground truths.

\paragraph{Multi-view Diffusion Models.}
Instead of text-to-image diffusion priors, some works learn image-conditioned novel view diffusion priors, which leverage the identity and appearance details of the input image to generate consistent novel views.  3DiM~\cite{3DiM} pioneered a diffusion model for pose-conditioned novel view generation.  Zero-123 finetuned StableDiffusion~\cite{LatentDiffusionModels} on the Objaverse~\cite{Objaverse, Objaverse-xl} dataset to improve generalization ability. Some works jointly diffused multiple views from a single image via dense 3D self-attention or epipolar attention layers in the denoiser network, including MVDiffusion~\cite{MVDiffusion},  SyncDreamer~\cite{Syncdreamer}, Zero-123++~\cite{Zero123++}, Wonder3D~\cite{Wonder3d}, MVDream~\cite{MvDream}, ImageDream~\cite{ImageDream}, Human3Diffusion~\cite{Human3Diffusion}, IM3D~\cite{Im-3d}, CAT3D~\cite{Cat3d}, and VideoMV~\cite{Videomv}. 
These works primarily target general objects, but our focus is on human heads. We train multi-view diffusion models on a multi-view head video dataset to obtain more head-specific priors. By using normal maps from FLAME reconstruction as the camera pose condition, we introduce pixel-aligned inductive bias to enable more precise viewpoint control, which is crucial for consistent head Gaussian supervision.

\begin{figure*}[h]
    \vspace{-6mm}
    \centering
 \includegraphics[width=\linewidth]
 {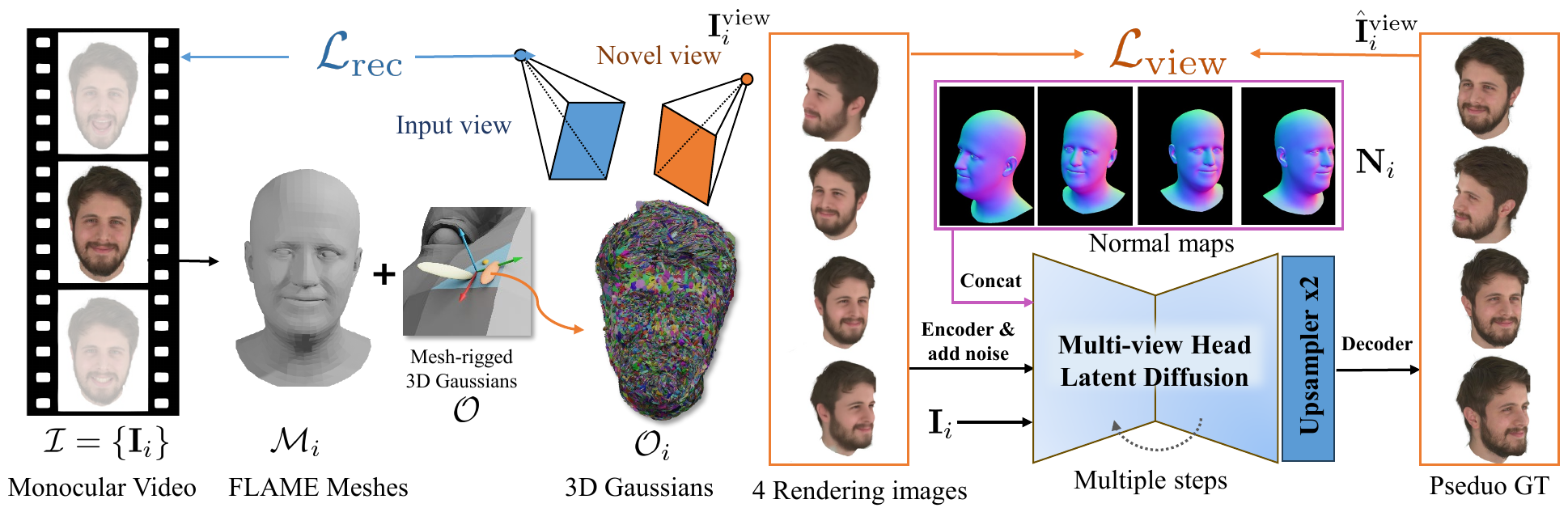}
   \vspace{-6mm}
  \caption{\textbf{Pipeline overview.} Given a sequence of RGB images from monocular cameras $\set{I}=\{ \Vec{I}_i \}$, our objective is to reconstruct dynamic head avatars by optimizing an animatable Gaussian splatting representation $\set{O}$, which is deformed to each frame as $\set{O}_i$ by the tracked FLAME mesh $\set{M}_i$ of $\Vec{I}_i$. We optimize $\set{O}$ by minimizing an input view reconstruction loss $\set{L}_{\rec}$, plus a view sampling loss $\set{L}_{\view}$. $\set{L}_{\view}$ compares novel-view renderings of $\set{O}_i$ from four random viewpoints $\Vec{I}_i^{\view}$, with pseudo ground truths $\Vec{\hat{I}}_i^{\view}$, predicted by a multi-view head latent diffusion model.   $\Vec{\hat{I}}_i^{\view}$ are generated by iteratively denoising 4-view latents, guided by the input image $\Vec{I}_i$ and normal maps $\Vec{N}_i$ rendered from $\set{M}_i$.  A latent upsampler module enhances facial details before decoding the denoised latent into an RGB image.
  }
    \label{fig:overview}
      \vspace{-3mm}
\end{figure*}
\section{Preliminaries}
\label{SecPreli}
\subsection{3D Gaussian Splatting}
3D Gaussian Splatting~\cite{GaussianSplatting} parameterizes a scene using a set of discrete geometric primitives known as 3D Gaussian splats. Each splat is characterized by a covariance matrix $\Sigma$ centered at location $\mu$.
The covariance matrices are required to be semi-definite for physical interpretations. It utilizes a parametric ellipsoid definition to construct the covariance matrix $\Sigma = RSS^TR^T$ using a scaling matrix $S$ and a rotation matrix $R$. These matrices are independently optimized and represented by a scaling vector $\Vec{s} \in \mathbb{R}^3$ and a quaternion $\Vec{q} \in \mathbb{R}^4$. $\Vec{r} \in \mathbb{R}^{3\times3}$ denotes the corresponding rotation matrix of $\Vec{q}$. To represent scene appearance,  sphere harmonic coefficients are used to represent the color $\Vec{c}$ of each Gaussian.
During rendering, a tile-based rasterizer is used to $\alpha$-blend all 3D Gaussians overlapping the pixel within a tile. To respect visibility order and avoid per-pixel sorting expenses, splats are sorted based on depth values within each tile before blending.

\subsection{Gaussian Avatars}
To animate Gaussian splats for head avatars, we use the representation of GaussianAvatars~\cite{GaussianAvatars} to rig 3D Gaussians using the FLAME mesh.
Initially, they bind each triangle of the FLAME identity mesh with a 3D Gaussian and transform the 3D Gaussian based on triangle deformations across different time steps. The splat remains static in the local space of the attached triangle but can dynamically evolve in the global metric space along rotation, translation, and scaling transformations of the binding triangle.
For each triangle, they compute the mean position $\Vec{T}$ of its three vertices coordinates as the origin of the local space. They define a rotation matrix $\Vec{R}$ to depict the orientation of the triangle in the global space, composed of three column vectors derived from the direction vector of one edge, the normal vector of the triangle face, and their cross product. Additionally, they determine triangle scaling $k$ by calculating the mean length of one edge and its perpendicular in a triangle.
The 3D Gaussian is parameterized by the location $\Vec{\mu}$, rotation $\Vec{r}$, and anisotropic scaling $\Vec{s}$ in the local space of its parent triangle. In the initialization stage, the location $\Vec{\mu}$ is set as zero, rotation $\Vec{r}$ as an identity matrix, and scaling $\Vec{s}$ as a unit vector. During rendering, they transform these properties from local to global space by:
\begin{equation}
\vspace{-2mm}
    \Vec{r}' = \Vec{R}\Vec{r}
\end{equation}
\begin{equation}
\vspace{-1mm}
    \Vec{\mu}' = k\Vec{R}\Vec{\mu} + \Vec{T}
\end{equation}
\begin{equation}
\vspace{-2mm}
    \Vec{s}' = k\Vec{s}
\end{equation}
The adaptive Gaussian densification process operates within local space, where newly added Gaussians inherit binding relations from their original counterparts. 
%
\section{Gaussian Avatars Fusion}
\label{SecGAF}
Given a monocular sequence of RGB images $\set{I}=\{ \Vec{I}_i \}$ as input, our goal is to reconstruct head Gaussian splats $\set{O}=\{ \set{O}_i \}$ as output. The overall pipeline is shown in Figure~\ref{fig:overview}. To enable animation of the reconstructed avatar with various head poses and expressions, we optimize the Gaussian splats $\set{O}$ rigged by a parametric head model,~\eg FLAME.  The challenge is that monocular videos often lack complete observations of the head. For example, a front-facing video with limited head rotations may provide insufficient information about the face’s sides. This poses difficulties for reconstructing a photorealistic 3D head from partial input, as the optimization over $\set{O}$ is underconstrained.
To address this, we introduce a normal map-guided, multi-view head diffusion model, which is designed to jointly denoise multi-view images conditioned on a single input image and normal maps rendered from FLAME tracking. Once trained, this model serves as a prior to regularize renderings from $\set{O}$, filling in unobserved regions and improving the quality of the reconstructed avatar.
%
%
%
\subsection{Normal map-conditioned Multi-view Head Latent Diffusion}
\begin{figure}[t]
\vspace{-4mm}
    \centering  \includegraphics[width=1.02\linewidth]{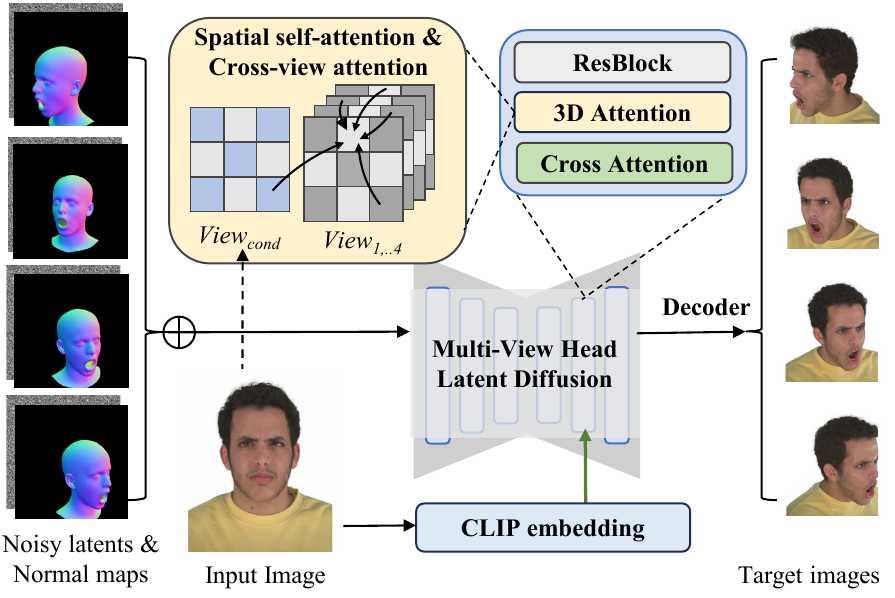}
    \vspace{-6mm}
  \caption{ \textbf{Multi-view latent head diffusion models.}
Given multi-view noisy image latents, we concatenate them with VAE latents of normal maps rendered from FLAME tracking. These combined inputs are processed by a 2D U-Net denoiser with attention blocks. To maintain 3D consistency, 3D attention blocks apply cross-attention across all views, integrating face identity and appearance details from the input image into the denoised latents while exchanging information between noisy latents across views. 
  }
    \label{fig:mvldm}
    \vspace{-4mm}
\end{figure}

To address missing regions in monocular videos with limited head coverage, one solution is to distill pre-trained text-to-image diffusion models to regularize novel view renderings. 
Personalized techniques like Dreambooth~\cite{Dreambooth} allow for identity preservation by customizing diffusion models for specific objects we wish to reconstruct.  
%
However, personalized text-to-image diffusion models are not designed to capture the distribution of novel views from a single input image. To overcome this, we propose a novel view diffusion model that generates identity-preserved and appearance-coherent novel views conditioned on an input image. By denoising multiple novel views simultaneously, our approach enhances cross-view consistency.
The multi-view head diffusion model is illustrated in Fig.~\ref{fig:mvldm}.

\paragraph{Normal Map Conditioning.} 
To control viewpoint, we leverage normal maps rendered from the FLAME mesh reconstruction at target views as diffusion guidance. The normal map is first encoded into a latent representation via pre-trained VAE~\cite{VAE},  matching the dimensionality of the noisy image latent, and these two latents are then concatenated along the channel dimension. Compared to camera pose embeddings, normal maps offer a more explicit inductive bias for view synthesis by providing pixel-aligned conditioning, which facilitates alignment between the generated images and the conditioning normal maps. Moreover, the FLAME renderings also facilitate the expression accuracy of synthesized novel views.

\paragraph{Model Architecture.} We train a multi-view diffusion model that takes a single image of a head, $\Vec{I}_{cond}$, as input and generates multiple output images conditioned on normal maps rendered from desired camera poses using the FLAME reconstruction $\set{M}$. Specifically, given $\Vec{I}_{cond}$ and the FLAME mesh $\set{M}$, the model learns the joint distribution of $N$ target images, $\Vec{I}_{tgt}$, guided by $N$ normal maps, $\Vec{N}_{tgt}$, which are rendered from $\set{M}$ at the target camera poses. 
\begin{equation}
\vspace{-3mm} 
    p(\Vec{I}_{tgt} | \Vec{I}_{cond}, \Vec{N}_{tgt} )
\end{equation}
 Our model architecture is similar to multi-view diffusion models (MVLDM)~\cite{MvDream,  ImageDream,  Human3Diffusion, Cat3d}, which is based on 2D U-Net~\cite{Unet} and attention blocks~\cite{Attention}.
 As shown in Fig.~\ref{fig:mvldm}, we use the CLIP image embedding to achieve global control over novel view generation. 
 However, this embedding mainly contains high-level semantic features, lacking the detailed information necessary for accurately capturing the head's identity and appearance. To address this, we incorporate the VAE latents of $\Vec{I}_{cond}$ directly into the multi-view diffusion model. By applying cross-attention between the multi-view denoised latents and the VAE latent of $\Vec{I}_{cond}$, we effectively transfer identity-specific details. Additionally, to ensure consistency across views, we aggregate information from the noisy latents across different viewpoints.

\subsection{Gaussian Avatars Reconstruction with Multi-view Diffusion Priors}
We now seek to utilize the multi-view diffusion priors for head Gaussian reconstruction. 
A commonly used strategy is Score Distillation Sampling (SDS) loss~\cite{Dreamfusion}, which performs one-step denoising.
Due to the stochastic nature of the denoising process introduced by random noise levels and seeds, it contains noisy gradients that disturb 3D optimization. Consequently, it often causes over-saturated appearance issues in synthesized 3D assets.
Wu et al.~\cite{ReconFusion} found that after iteratively denoising a noisy image for multiple steps, we can obtain a deterministic output. 

\paragraph{Pseudo-image Ground Truths.} 
At each iteration, we randomly select the $i$-th input frame $\Vec{I}_i$ and its FLAME mesh $\set{M}_i$.
By randomly sampling 4 viewpoints $\{\phi_j\}_{j=1}^{4}$, then we can generate 4 novel views from $\Vec{I}_i^{\view}= \set{R}(\set{O}_i, \{\phi_j\} )$, and render normal maps $\vec{N}_i = \set{R}(\set{M}, \{\phi_j\})$  from  $\set{M}_i$.
We can employ normal map-guided multi-view diffusion priors to regularize $\Vec{I}_i^{\view}$. 
Concretely, we encode $\Vec{I}_i^{\view}$ into latent features $\Vec{z}$, which is perturbed with noise $\epsilon_t$ to obtain noisy features $\Vec{z}_t$.  The noise intensity of $\epsilon_t$ is controlled by the diffusion time step $t \sim [ 0.02, 0.98]$.
We then iteratively perform multiple denoising steps of latent diffusion until the final clean latent $\Vec{z}_0$ is obtained.
To accelerate the generation speed, we adopt the DDIM sampling strategy~\cite{DDIM} by running $t/k$ intermediate steps, which can reduce the denoising steps by $k=20$  times.
Then $\Vec{z}_0$ is decoded back to 4 images $\{ \Vec{\hat{I}}_i^{j} \}$ that are served as pseudo supervision.
%

\paragraph{Latent Upsampler.}
To further enhance the facial details in $\Vec{\hat{I}}_i^{\view}$, we use a pre-trained latent upsampler diffusion model~\cite{sdx2latentupsampler} to super-resolve the denoised 4-view latents $\Vec{z}_0$, from a resolution of 32 $\times$ 32 to 64 $\times$ 64. This super-resolution step allows the pseudo-image ground truths $\Vec{\hat{I}}_i^{j}$  to reach a final resolution of 512$\times$512. We use 10-step DDIM sampling in our latent upsampler inference. 

\paragraph{3D-Aware Denoising.}
The rendered views $\Vec{I}_i^{\view}$ from a global 3D representation $\Vec{O}_i$ introduce 3D-awareness into the denoising process, further enhancing multi-view consistency across pseudo ground truths $\Vec{\hat{I}}_i^{j}$.

\subsection{Loss Functions}
We supervised the optimization of Gaussian Splats $\set{O}$ by a combination of loss functions in the following:
\begin{equation}
\label{EquaAllLoss}
\vspace{-2mm}
\begin{aligned}
     \set{L} = \set{L}_{\img}(\Vec{I}^{\rec}, \Vec{I}) + L_{\img}(\Vec{I}^{\view}, \Vec{\hat{I}}^{\view})   \\ 
    + \lambda_{\pos} \set{L}_{\pos} + \lambda_{\scale} \set{L}_{\scale}
\end{aligned}
\end{equation}
$L_{\img}$ is defined by a combination of pixel-wise $L_1$ loss, SSIM loss, and LPIPS loss:
\begin{equation}
\vspace{-2mm}
    \set{L}_{\img} = \lambda_1 \set{L}_1 + \lambda_2 \set{L}_{SSIM} + \lambda_3 \set{L}_{LPIPS} 
\end{equation}
where $\lambda_1 = 0.8$,  $\lambda_2 = 0.2$, and $\lambda_3 = 0.1$.
We also introduce splat position and scale regularization terms to penalize abnormally distributed splats.  $\lambda_{\pos}$ and $\lambda_{\scale} $ are set to 0.01 and 1 respectively.

\subsection{Implementation Details}
\paragraph{\bf{Multi-view Head Diffusion Model. }}
It is initialized from Stable Diffusion 2.1~\cite{LatentDiffusionModels} of ImageDream~\cite{ImageDream} and is trained on eight A100 GPUs over 20,000 iterations, taking approximately 72 hours. The training uses a learning rate of 0.0001 and a batch size of 64. During training, we employ a classifier-free guidance strategy, randomly dropping the input image at a rate of 0.1. The model is trained on the multi-view human head video dataset NeRSemble~\cite{Nersemble}, which contains RGB video sequences from 16 viewpoints, covering both front and side faces. We randomly sample 50,000 timesteps to construct the training dataset.

\paragraph{\bf{Gaussian Avatar Optimization.}}
The FLAME meshes are initially obtained by VHAP tracker~\cite{qian2024versatile} from monocular videos. The animatable Gaussians are optimized with Adam~\cite{kingma2014adam} for 6,000 iterations, with learning rates of $\expnumber{5}{-5}$,$\expnumber{1.7}{-2}$, $\expnumber{1}{-3}$, $\expnumber{2.5}{-3}$, and $\expnumber{5}{-2}$ for splat position, scaling factor, rotation quaternion, color and opacity respectively. We perform adaptive densification if the position gradients are larger than $0.0002$ in every 300 iterations until 5,000 iterations are reached. We also remove Gaussians with opacity less than 0.005. During training, we also finetune the FLAME parameters using the learning rates $\expnumber{1}{-6}$, $\expnumber{1}{-5}$, and $\expnumber{1}{-3}$ for translation, joint rotation, and expression coefficients. 
{
 At each iteration, we randomly sample 4 out of 15 novel viewpoints from the NeRSemble dataset to calculate multi-view diffusion loss.
 }
%
 %

\paragraph{\bf{Runtime.}}
Our current implementation of avatar reconstruction takes about 12 hours and uses 32 GB of memory on a single A6000 GPU. 
After avatar reconstruction, { Our method takes 0.016 seconds to render an image at 802$\times$550 resolution, i.e. 62 FPS.}  
 
\section{Experiments}\label{SecExp}
\paragraph{\bf{Datasets.}}
We conduct head avatar reconstruction experiments on monocular video sequences from the \textbf{NeRSemble dataset}~\cite{Nersemble}. Note that these evaluation sequences were not seen by the multi-view diffusion model. 
We use monocular videos from the 8-th camera as the input, only capturing the head from the front view. And we use videos from the other 15 views for evaluation.  We randomly select 12 sequences from different identities, with durations between 70 and 300 frames, downsampled to a resolution of $802 \times 550$. 
Additionally, we include a \textbf{Monocular Video} dataset consisting of 3 monocular videos from the INSTA~\cite{INSTA} dataset at 
$512 \times 512$ resolution, and 3 sequences of three subjects captured by smartphone at 
$1280 \times 720$ resolution. 
%
%

\paragraph{\bf{Evaluations.}}
Following previous works~\cite{INSTA, GaussianAvatars}, we report the average L1 loss, LIPIS, PSNR, and SSIM between renderings and ground truths of the test set.
In the NeRSemble dataset, we evaluate head avatar reconstruction and animation quality in two settings:
1) \emph{novel view synthesis}: driving a reconstructed avatar with seen head poses and expressions during training, and rendering it from 15 hold-out viewpoints. 
2) \emph{novel expression synthesis}:   driving a reconstructed avatar with unseen head poses and expressions during training, and rendering it, rendered from 5 nearby hold-out views,~\ie cameras 6–10. 
For this dataset, 80\% of frames are used for training, and 20\% for evaluation.
%
In the Monocular Video dataset, we use around 40\% of frames for training and the rest for evaluation. The training sets consist of front-facing frames, while the evaluation set includes obvious head rotations to capture side faces.

\paragraph{\bf{Baselines.}}
{
We compare our method against state-of-the-art methods for photo-realistic head reconstruction.
\emph{PanoHead}~\cite{Panohead} and \emph{SphereHead}~\cite{Spherehead}  use 3D-GAN inversion to reconstruct static heads.  They cannot reconstruct animatable avatars.  Thus, they cannot be evaluated in novel expression synthesis.
\emph{INSTA}\cite{INSTA}, \emph{FlashAvatar}~\cite{FlashAvatar}, and \emph{GA}\cite{GaussianAvatars} optimize photo-realistic head avatars represented by Instant NGP or Gaussian Splatting.  
They overfit avatars to monocular observations.
\emph{P4D-v2}~\cite{Portrait4Dv2} and \emph{GAGAvatar}~\cite{GAGAvatar} reconstruct animatable 3D avatars from single images, that can then be reenacted with novel expressions. 
}

%
\subsection{Head Avatar Reconstruction}
\begin{table}[t]
	\renewcommand\arraystretch{1.1}
        \setlength{\tabcolsep}{1.8pt}
	\begin{center}
        \footnotesize
		\begin{tabular}{*{7}{c}}
                \toprule
                \multirow{2}*{Method} & \multicolumn{3}{c}{Novel Views}   & \multicolumn{3}{c}{Novel Expressions}  \\
			\cmidrule(lr){2-4} \cmidrule(lr){5-7} 
			    & LPIPS $\downarrow$  & PSNR $\uparrow$ & SSIM $\uparrow$  
                    & LPIPS $\downarrow$  & PSNR $\uparrow$ & SSIM $\uparrow$   \\ 
                \midrule
                \midrule
               PanoHead~\cite{Panohead}  &    0.171   &  17.40  &  \tabsecond{86.11}  & N/A & N/A & N/A  \\
                SphereHead~\cite{Spherehead}  &   0.174    &  16.94  &  85.98  & N/A & N/A & N/A  \\
                \midrule
                INSTA~\cite{INSTA}  & 0.262   & 15.87 & 77.02 & 0.165  & 19.46  & 84.91  \\
                FlashAvatar~\cite{FlashAvatar} & 0.247 & 16.94  & 81.05  & 0.145  & 21.37 & 86.08  \\
                GA~\cite{GaussianAvatars}          & 0.218  & 17.51  & 81.66 & 0.138  &  21.63  & 87.00  \\
                P4D-v2~\cite{Portrait4Dv2}    &  \tabsecond{0.161}   &  16.82  &  85.44  & \tabsecond{0.142} & 18.72 & \tabsecond{86.63}   \\ 
                 GAGAvatar~\cite{GAGAvatar}    & 0.183  & \tabfirst{21.20}  & 80.72  &  0.177 & \tabsecond{21.32}  &  80.44   \\ 
                \midrule
                 Ours     & \tabfirst{0.125} &  \tabsecond{20.88}  &  \tabfirst{88.91} &  \tabfirst{0.087}  &   \tabfirst{24.12} &   \tabfirst{90.66}  \\
                \bottomrule
        \end{tabular}
        \vspace{-2mm}
        \caption{Quantitative comparisons on dynamic avatar reconstruction and animation from monocular videos. Results are obtained by the average of twelve sequences of different subjects on the \textbf{Nersemble dataset}. ~\tabfirst{Best} and~\tabsecond{2nd-best} are highlighted.
        }
        \label{tab:compare_sota}
        \vspace{-4mm}
        \end{center}
\end{table}

%
\begin{figure*}[t]
    \vspace{-6mm}
    \centering
    \includegraphics[width=\linewidth]{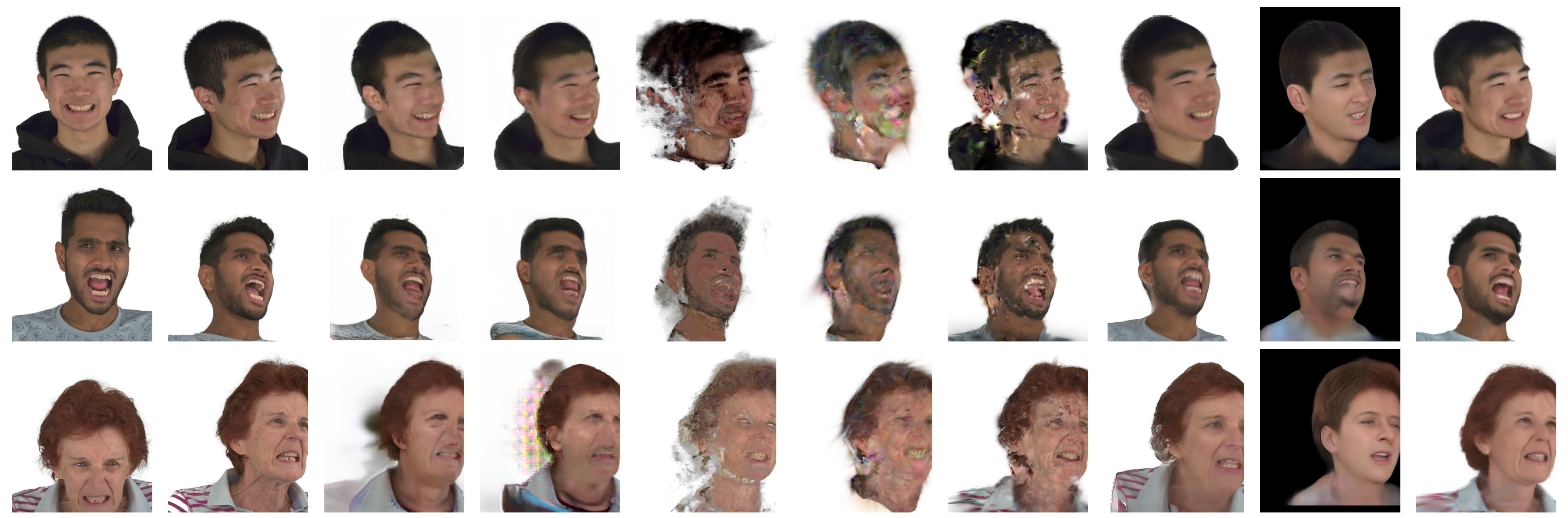}
    \vspace{-2mm}
    \begin{tabularx}{\textwidth}{XXXXXXXXXX}
        \centering \small Input & \centering \small GT   & \centering \small PanoHead & \centering \small SphereHead & \centering \small INSTA & \centering \small FlashAvatar 
        & \centering \small GA & \centering \small P4D-v2 & \centering \small GAGAvatar & \centering \small Ours
    \end{tabularx}
    \caption{\textbf{Novel view synthesis from monocular videos on the NeRSemble dataset.}  { Compared to state-of-the-art methods, our approach reconstructs unseen side facial regions in the inputs, better preserves facial identities, and consistently produces more favorable renderings from hold-out views. }
    } 
    \label{fig:novelview}    
\end{figure*}
\begin{figure}[h]
     \vspace{-2mm}
    \centering
    \includegraphics[width=\linewidth]{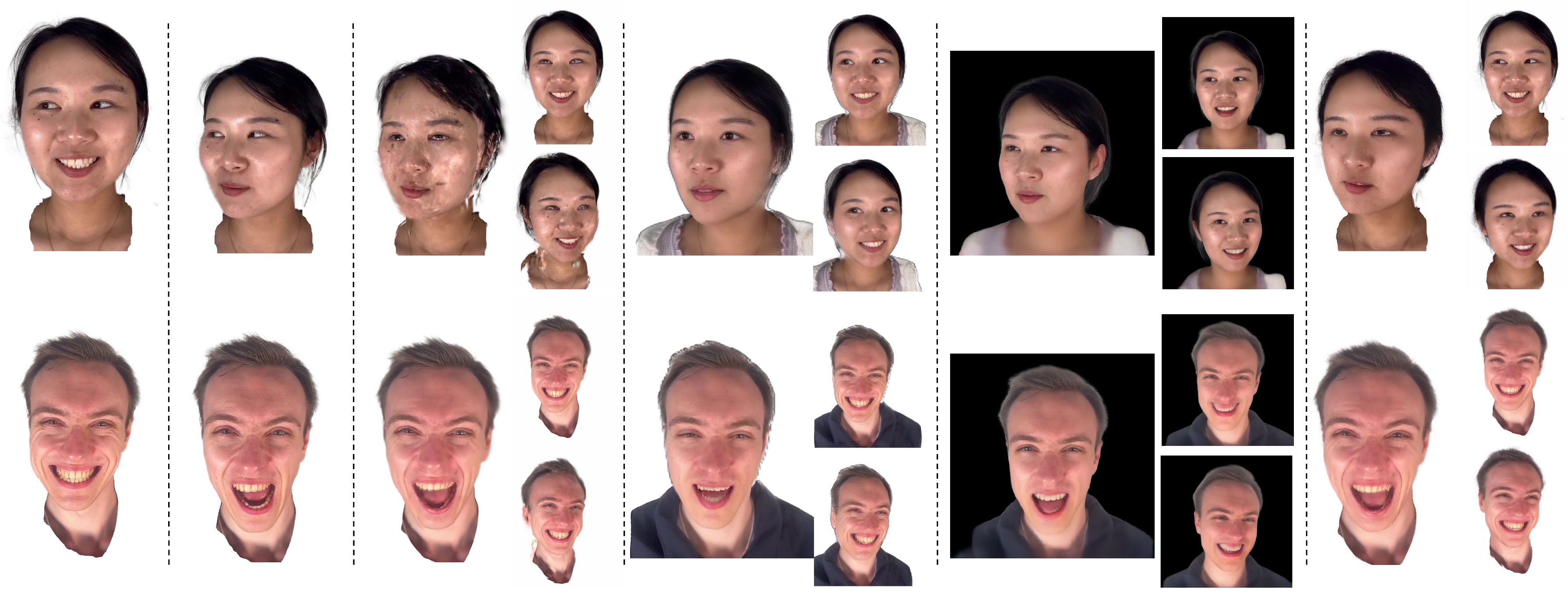}
    \begin{tabular*}{\linewidth}{@{\extracolsep{\fill}}p{0.08\linewidth}p{0.08\linewidth}p{0.10\linewidth}p{0.15\linewidth}p{0.12\linewidth}p{0.12\linewidth}}
    \centering \small Input & \centering \small (a)  & \centering \small (b) & \centering \small (c) & \centering \small (d) & \centering \small (e)  \\
    \end{tabular*}
    \vspace{-6mm}
    \caption{
    \textbf{Head avatar reconstruction from monocular videos captured on commodity devices.}  (a) Ground truth of novel expressions; (b) GA; (c) P4D-v2; (d) GAGAvatar; (e) Ours. 
    For each method, we display the fitting results of the input frame (top right) and novel view renderings of the input frame (bottom right). 
    {
    Given a front-facing sequence with limited head poses,  all methods can accurately reconstruct the observed frames. However, without effective priors to constrain unobserved regions, GA struggles to generalize to novel views and poses.  P4D-v2 and GAGAvatar exhibit limitations in capturing complicated facial expressions or fine-grained details such as wrinkles.}
    }
    \vspace{-4mm}
    \label{fig:novelview_real}
\end{figure}
%
\paragraph{NeRSemble.} 
As shown in Fig.~\ref{fig:novelview}, front-facing monocular videos lack sufficient side-face information. INSTA, FlashAvatar, and GA can only reconstruct observed regions and leave unobserved areas unconstrained. This often leads to artifacts in extreme hold-out views. 
{PanoHead~\cite{Panohead} and SphereHead~\cite{Spherehead} fail to generate plausible renderings in extreme novel views.
P4D-v2~\cite{Portrait4Dv2} exhibits blurriness and artifacts in the mouth and ears, while GAGAvatar~\cite{GAGAvatar} suffers from identity drift due to weak constraints in less observed regions.
In contrast, our method leverages multi-view diffusion priors to improve novel view synthesis, effectively completing missing regions and enhancing photorealism while preserving identity and appearance consistency. As shown in Tab.~\ref{tab:compare_sota}, our approach outperforms all baselines in LPIPS and SSIM and ranks second in PSNR on the task of novel view synthesis. Interestingly, novel view constraints also benefit novel expression synethesis, resulting in superior performance across all metrics.} 

\paragraph{Monocular Videos.}
\begin{table}[t]
        \vspace{-2mm}
	\renewcommand\arraystretch{1.1}
        \setlength{\tabcolsep}{3.2pt}
	\begin{center}
        \footnotesize
		\begin{tabular}{*{5}{c}}
                \toprule
			    & L1 $\downarrow$ & LPIPS $\downarrow$  & PSNR $\uparrow$ & SSIM $\uparrow$  
                     \\ 
                \midrule
                INSTA~\cite{INSTA}  &  0.0407  &  0.177  &  21.11  &  85.04   \\
                FlashAvatar~\cite{FlashAvatar} &  0.0376  &  0.145 &  21.13 &  85.93   \\
                GA~\cite{GaussianAvatars}  &  0.0263  &  0.109  & 22.55   &   88.61   \\
                P4D-v2~\cite{Portrait4D}  & 0.0317 & 0.131 & 20.95 & 86.82 \\
                GAGAvatar~\cite{GAGAvatar}  & 0.0282  & 0.108  & 22.34 & 85.38 \\
                Ours    &  \bf 0.0229  & \bf 0.090  & \bf 23.16  &  \bf 89.76  \\
                \bottomrule
        \end{tabular}
        \vspace{-2mm}
        \caption{Quantitative comparisons on dynamic avatar reconstruction on \textbf{Monocular Video dataset}. Results are obtained by \textbf{hold-out expression frames} from six sequences of INSTA and smartphone capture. Note that \textbf{novel view renderings cannot be evaluated} due to the absence of ground truths in single-view captures.}
        \label{tab:compare_mono}
        \vspace{-6mm}
        \end{center}
\end{table}
We also provide the comparisons on the Monocular Video dataset. Since single-view datasets lack ground-truth novel views, we evaluate animation results by applying expression parameters from hold-out frames. As shown in Tab.~\ref{tab:compare_mono}
, our method consistently surpasses others in all metrics. The qualitative results are presented in Fig.~\ref{fig:novelview_real}. 
{
While all methods can well reconstruct the input view (top right), GA~\cite{GaussianAvatars} struggles to generate plausible novel views (bottom right), particularly for unseen face regions. It is also prone to artifacts and inconsistencies in novel pose animations.
P4D-v2~\cite{Portrait4Dv2} and GAGAvatar~\cite{GAGAvatar} have limited capacity to represent exaggerated expressions or fine details such as wrinkles.
}

\subsection{Ablation Studies}
\begin{figure*}[!htp]
     \vspace{-4mm}
    \centering
    \includegraphics[width=\linewidth]{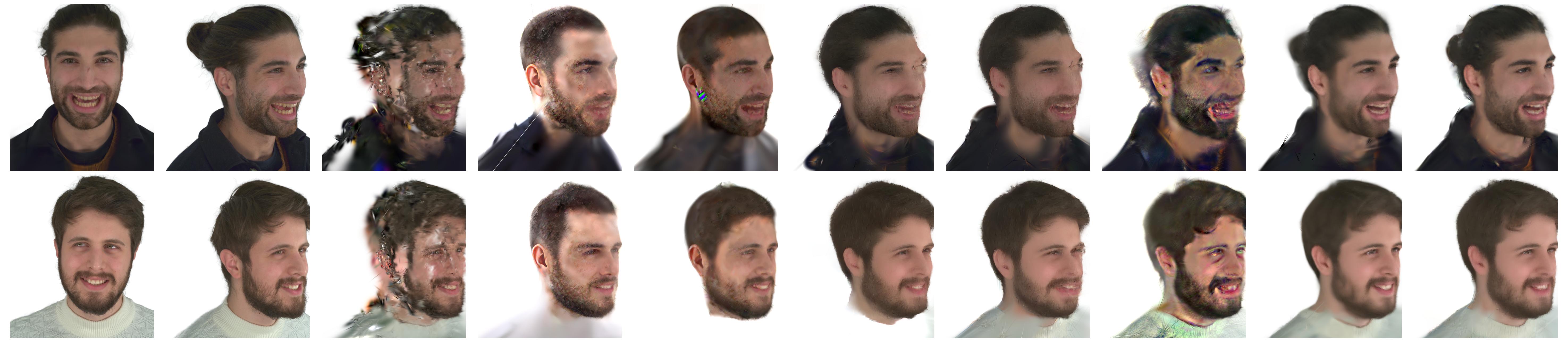}
    \begin{tabular*}
        {\linewidth}{@{\extracolsep{\fill}}p{0.001\linewidth}p{0.07\linewidth}p{0.07\linewidth}p{0.07\linewidth}p{0.07\linewidth}p{0.07\linewidth}p{0.07\linewidth}p{0.07\linewidth}p{0.07\linewidth}p{0.07\linewidth}p{0.07\linewidth}}
           & \centering Input  & \centering GT & \centering (a) & \centering (b) & \centering (c)  & \centering (d) & \centering (e) & \centering (f) & \centering (g) & \centering (h) 
    \end{tabular*}
     \vspace{-5mm}
    \caption{
        \textbf{Ablation studies on different types of diffusion priors.}  Comparisons between method variants of (a) No diffusion; using (b)  Pretrained Stable Diffusion; (c) Personalized Stable Diffusion; (d) Pose-conditioned multi-view diffusion; (e) Raymap-conditioned multi-view diffusion; 
        (f) Our multi-view diffusion using Score Distillation Sampling (SDS) loss; (g) Ours without latent upsampler $\times$2; (h) Our final model.  Our normal map-conditioned multi-view diffusion priors enable more photo-realistic novel views with identity and appearance consistency, by constraining novel views using pseudo-image ground truths, which are  decoded from iteratively denoised latents followed by the latent upsampler.
    }
    \label{fig:ablation}
    \vspace{-3mm}
\end{figure*}
%
We conduct detailed ablation studies to evaluate diffusion prior choices for novel view constraints, confirming the effectiveness of each design in our face-specific multi-view diffusion prior learning for 3D head reconstruction from monocular videos.  We use six sequences from the NeRSemble dataset.
The results are presented in Fig.~\ref{fig:ablation}, and Tab.~\ref{tab:ablation}.
Please refer to the supplementary material for more implementation details of ablation studies. 
\begin{table}[t]
	\renewcommand\arraystretch{1.1}
    \setlength{\tabcolsep}{1.1pt}
	\begin{center}
        \footnotesize
        \begin{tabular}{*{7}{c}}
            \toprule
            \multirow{2}*{Method} & \multicolumn{3}{c}{Novel Views}   & \multicolumn{3}{c}{Novel Expressions}  \\
        \cmidrule(lr){2-4} \cmidrule(lr){5-7}
                 & LPIPS $\downarrow$  & PSNR $\uparrow$ & SSIM $\uparrow$   
                 & LPIPS $\downarrow$  & PSNR $\uparrow$ & SSIM $\uparrow$   \\ 
            \midrule
            \midrule
            No diffusion & 0.207 & 18.47 & 82.69 & 0.129 & 22.74 & 88.13 \\
            pretrained SD & 0.196 & 16.99 & 83.47 & 0.149 & 21.37 & 87.81 \\
            personalized SD & 0.196 & 16.89 & 85.98 & 0.147 & 21.51 & 89.52 \\
            Ours, w/o pseduo GT & 0.178 & 19.57 & 84.73 & 0.136 & 22.41 & 87.59 \\
            Ours, pose cond. & 0.131 & 19.69 & 87.63 & 0.087 & 23.68 & 90.37 \\
             Ours, raymap cond.   &  0.150 & 18.56 & 86.70 & 0.095  & 22.84 &  89.73 \\
            Ours, w/o latent$\times$2 & 0.134 & 21.48 & 89.53 & 0.087 & 25.07 & 91.75 \\
            \midrule
            Ours final & \bf 0.118 & \bf 21.82 & \bf 89.87 & \bf 0.079 & \bf 25.39 & \bf 92.02 \\
            \bottomrule
        \end{tabular}
        \vspace{-2mm}
        \caption{ \textbf{Ablation Studies on different types of diffusion priors.} Results are obtained from the average of six sequences of different subjects from the NeRSemble~\cite{Nersemble} dataset.  
        } 
        \label{tab:ablation}
        \end{center}
        \vspace{-6mm}
\end{table}

\paragraph{\bf{What is the effect of multi-view head diffusion priors?}}
An alternative is to use pre-trained text-to-image diffusion models Stable Diffusion~\cite{LatentDiffusionModels}. Another alternative is to fine-tune Stable Diffusion on RGB frames from the input video, obtaining personalized image diffusion priors. We implement both variants using normal map guidance of ControlNet~\cite{ControlNet}.  In Fig.~\ref{fig:ablation} (b)
, pre-trained text-to-image priors often produce renderings that deviate from the original identity.  In Fig.~\ref{fig:ablation} (c), personalized diffusion priors improve identity preservation but struggle with appearance consistency, as they lack input image information to hallucinate novel views. Our approach learns to jointly generate multiple novel views conditioned on the input image, thus achieving higher realism and view consistency in both identity and appearance.

%
\paragraph{\bf{What is the effect of normal map conditioning for multi-view diffusion models?}}
Our multi-view head diffusion models condition on normal maps of FLAME reconstruction, which are pixel-aligned with the target novel-view images. Fig.~\ref{fig:ablation} (d) and (e) reflect that using camera pose embedding and ray map conditioning could introduce obvious misalignment errors in synthesizing pseudo-image ground truths, leading to blurred 3D Gaussian renderings.

\paragraph{\bf{What is the effect of iteratively denoised images as pseudo-ground-truths?} }
%
We can instead use Score Distillation Sampling (SDS) loss to constrain multi-view renderings.  As shown in Fig.~\ref{fig:ablation} (f), vanilla one-step SDS loss has appearance oversaturation issues in the face region.

\paragraph{\bf{What is the effect of latent upsampler module?}}
Through Fig.~\ref{fig:ablation} (g) vs. (h), we can see that the upsampler module significantly sharpens and enhances facial appearance details.
%

\subsection{Multi-view Head Diffusion Models}
\begin{figure}[!htp]
\vspace{-1mm}
    \centering
    \includegraphics[width=.92\linewidth]{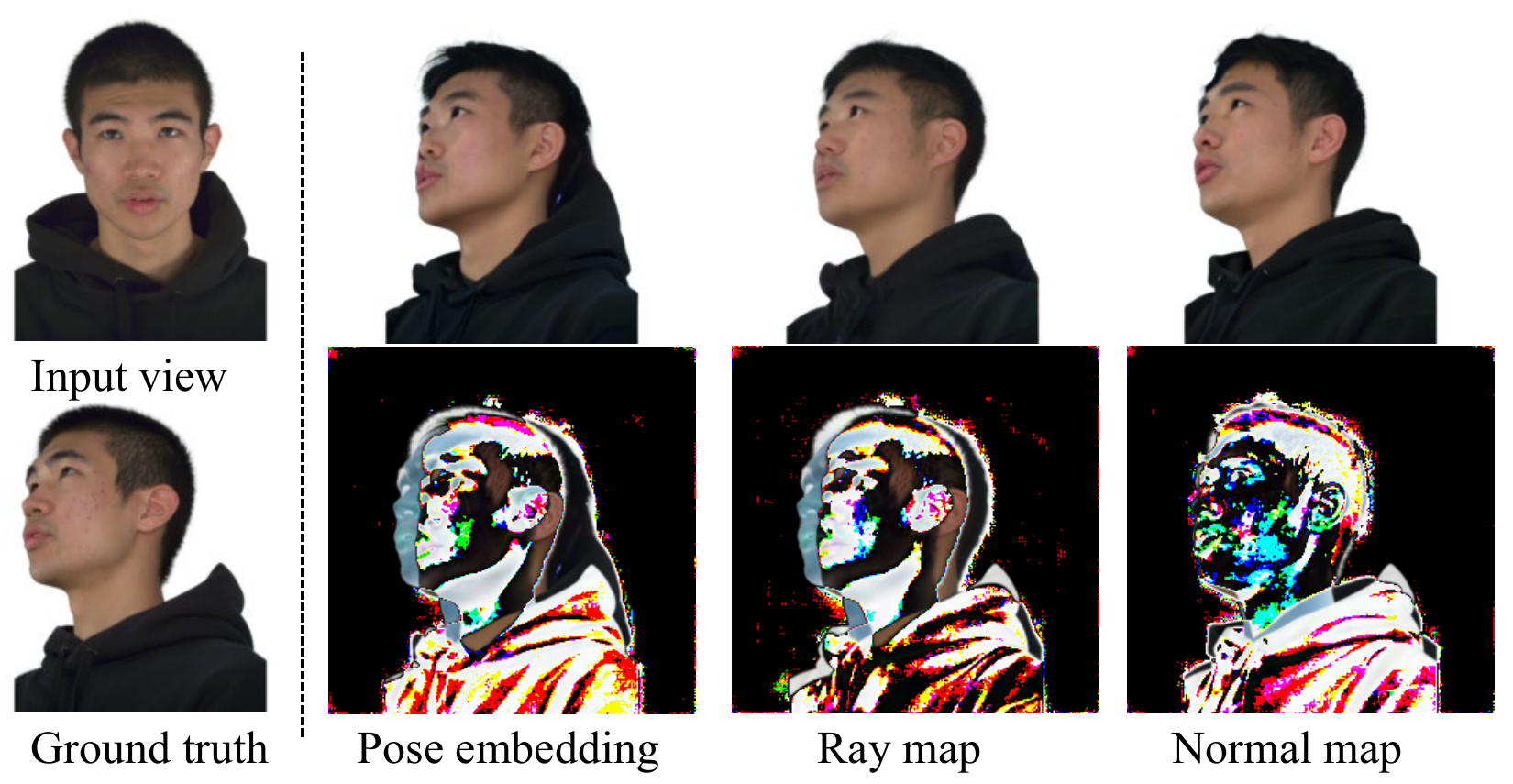}
    \vspace{-1.5mm}
    \caption{\textbf{Ablation studies on different camera conditioning in multi-view head diffusion.} 
    We show predictions in the first row and $l_1$ difference maps between the ground truth and predictions.}
    \label{fig:ablation_mvlhd}
\vspace{-3mm}
\end{figure}

{We compare normal map conditioning to alternatives in our multi-view head diffusion models, including pose embedding and Pl\"{u}cker ray maps~\cite{plucker1865xvii}. Results in Tab.~\ref{tab:ablation_mvlhd} and Fig.~\ref{fig:ablation_mvlhd} show that FLAME normal maps can achieve more superior facial alignment, with $\times2.56$ lower $l_1$ error and $7.73\%$ higher SSIM.}
\begin{table}[!htbp]
    \vspace{-2mm}
    \setlength{\tabcolsep}{1.5pt}
	\begin{center}
	\begin{tabular}{*{5}{c}}
                \toprule
			Method & $l_1$ $\downarrow$ & LPIPS $\downarrow$  & PSNR $\uparrow$ & SSIM $\uparrow$  \\ 
                \midrule
               pose embedding  &  0.123  & 0.184  & 16.52  & 76.12\\
               ray map      &  0.137  & 0.198  & 16.05 & 75.71 \\
               normal map   &  \bf 0.048  & \bf 0.119 & \bf 19.67  & \bf 83.85 \\
                \bottomrule
        \end{tabular}
       \vspace{-2mm}
        \caption{\textbf{Ablation studies on different camera conditioning in multi-view head diffusion.} }
        \label{tab:ablation_mvlhd}
        \end{center}
    \vspace{-8mm}
\end{table}
%

%
%
\subsection{Limitations and Future Work}
While our work has shown promising results in dynamic avatar reconstruction from monocular videos captured on studio setups or commodity devices, there are limitations in our current method.
First, we do not explicitly separate the material and appearance of heads, which could enable re-lighting applications~\cite{saito2024relightable, SwitchLight}.
Second, optimizing head Gaussians using iteratively updated pseudo ground-truths from diffusion models is time-consuming. We plan to explore real-time 4D avatar reconstruction with feed-forward large reconstruction models~\cite{LRM, GSLRM, Instant3d}.
Lastly, the quality of our avatar reconstruction and animation is limited by the expressiveness of current parametric head models, which lack detailed hair geometry and animation. Future work could extend Gaussian head avatars to incorporate fine-grained hair modeling and animation.~\cite{HairGaussians, GaussianHair}.


\section{Conclusion}
%
%
In this work, we present a novel method to reconstruct photo-realistic head avatars from monocular videos to push the frontier of avatar fidelity from commodity devices. 
%
Due to limited observation and coverage of human heads, Gaussian reconstruction from monocular videos is inherently under-constrained. To address this challenge, we introduce multi-view diffusion priors that jointly constrain photorealism across multiple views rendered from Gaussian splats.
{
We obtain these priors by designing a multi-view head diffusion model, fine-tuned on a multi-view head video dataset to generate novel views from a single image, conditioned on rendered normal maps from FLAME tracking. To mitigate over-saturated appearances, we use iteratively denoised images as pseudo-ground truths.
 A pre-trained latent upsampler diffusion model further enhances facial details.}
By reducing data capture requirements for avatar creation, our approach has the potential to unlock new opportunities in immersive VR/AR applications and products.

\section*{Acknowledgements}
This work was supported by Toyota Motor Europe and Woven by Toyota. This work was also supported by the
ERC Starting Grant Scan2CAD (804724). We thank Rundi Wu, Simon Giebenhain, and Lei Li for constructive discussions
and Angela Dai for the video voice-over.
{
    \small
    \bibliographystyle{ieeenat_fullname}
    \bibliography{main}
}

\clearpage
\appendix
\section*{Appendix}
In this supplementary material, we provide additional information about the dataset in Sec.~\ref{SecData}. Subsequently, we present more detailed explanations about method implementations in Sec.~\ref{SecImple}, including parametric head tracking and multi-view head diffusion.
Following that, we showcase the results of our multi-view head diffusion results in Sec~\ref{SecMVHDRes}.
Next, we provide additional comparisons in Sec~\ref{SecAddRes}, including novel view synthesis, self-/cross-reenactment, and robustness analysis.
Finally, we discuss the ethical considerations and potential negative impacts in Section~\ref{SecImpact}.
\section{Dataset}
\label{SecData}

\paragraph{Smartphone Video Capture.}
We capture monocular video sequences using an iPhone 14 Pro.
The subject is seated in a chair, and the room lights are turned on during the recording, providing adequate illumination. The duration of the recording is about 10-15 seconds, at 30 frames per second. The image resolution is 1280 $\times$ 720.

\begin{figure}[!htp]
    \centering
    \begin{subfigure}{0.31\linewidth}
        \includegraphics[width=\linewidth]{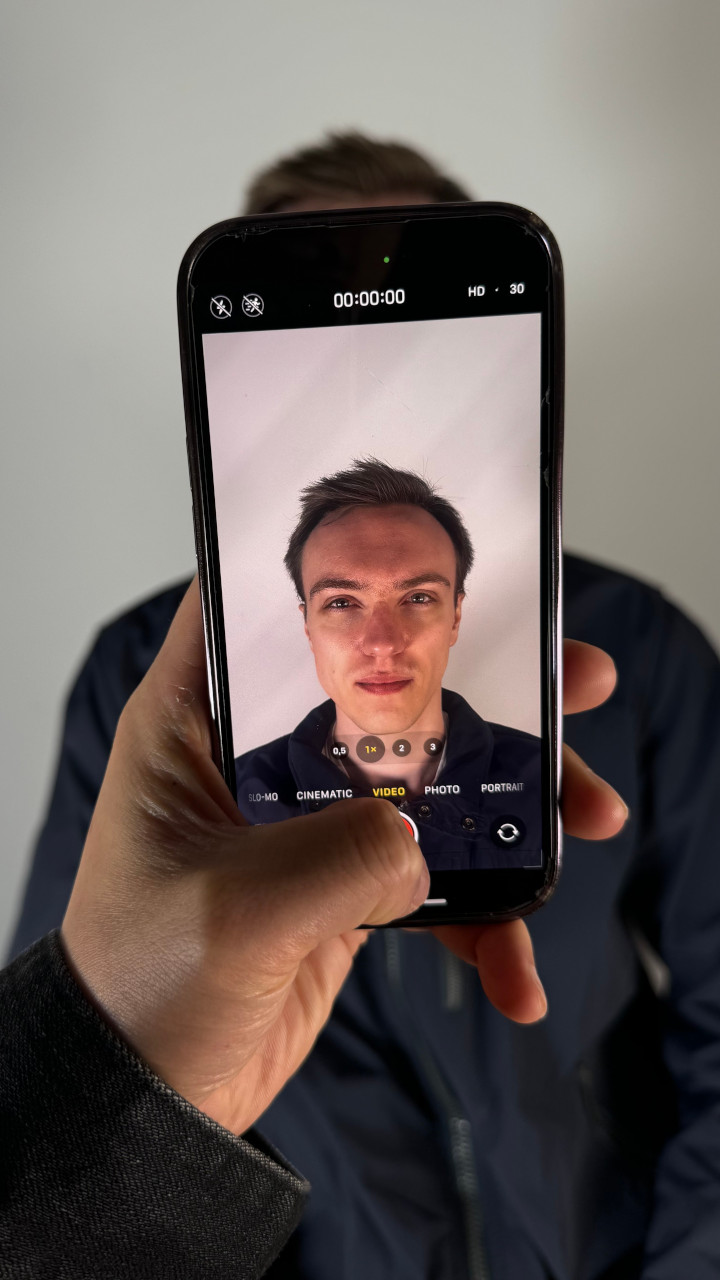}
        \caption{Video capture}
        \label{fig:image1}
    \end{subfigure}
    \hfill
    \begin{subfigure}{0.31\linewidth}
        \includegraphics[width=\linewidth]{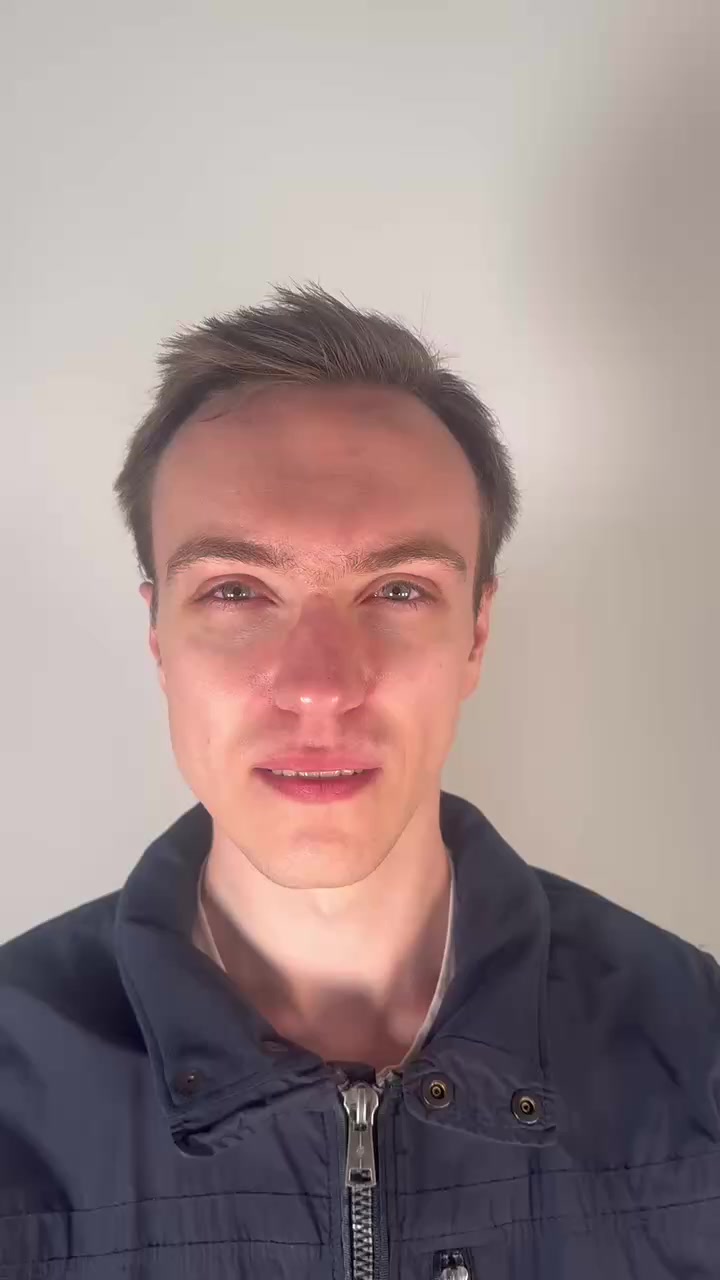}
        \caption{Raw image}
        \label{fig:image2}
    \end{subfigure}
    \hfill
    \begin{subfigure}{0.31\linewidth}
        \includegraphics[width=\linewidth]{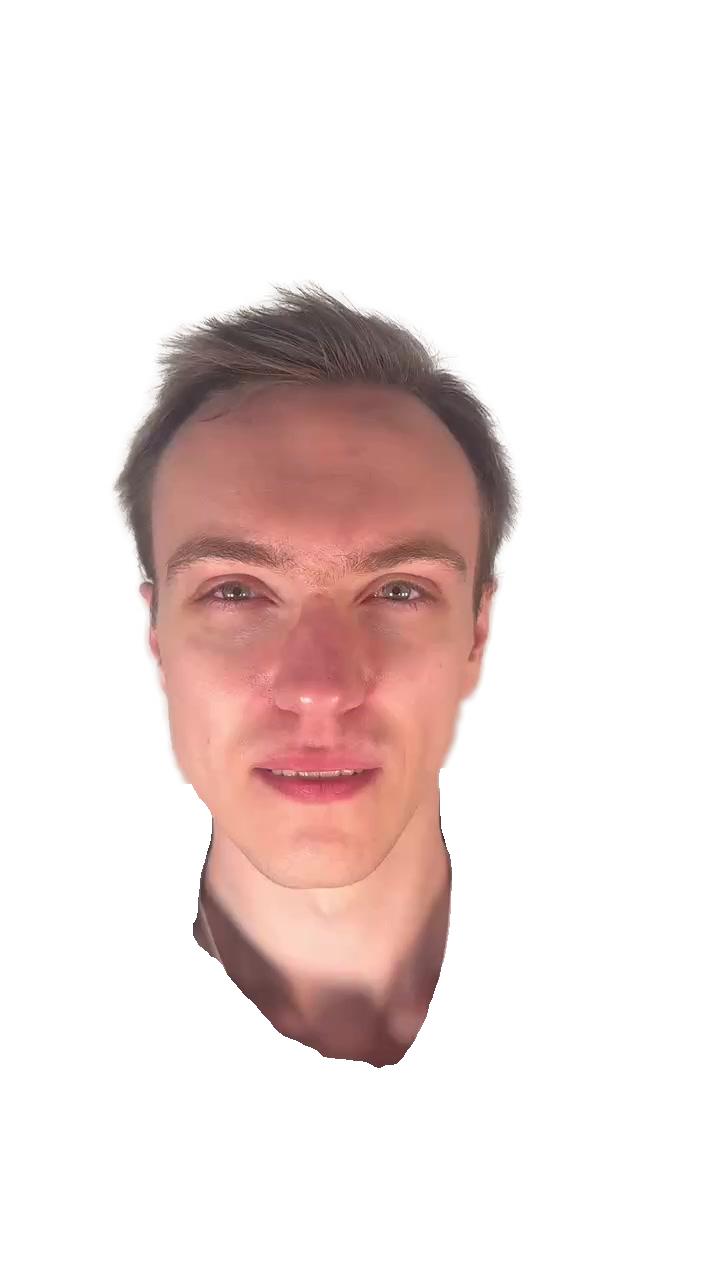}
        \caption{Processed image}
        \label{fig:image2}
    \end{subfigure}
    \caption{\textbf{Data capture and processing of monocular videos from smartphones.} We capture a short video using an iPhone 14 Pro. From the raw images, we remove the background using image matting techniques and segment out the torso to focus on the head region. }
    \label{fig:datapre}
\end{figure}

\paragraph{Data preprocessing.}
To simplify the optimization process for animatable Gaussian splats, we integrate two preprocessing steps on raw images extracted from monocular videos. Firstly, we leverage the image matting techniques proposed in \cite{BackgroundMat, rvm} to remove the background. More specifically, we use \cite{rvm} for our smartphone video capture, while we adopt \cite{BackgroundMat} for the NeRSemble \cite{Nersemble} dataset, where the initial background image is provided. Secondly, we utilize face segmentation maps acquired from BiSeNet~\cite{Bisenet} to isolate and crop out the torso portion, thus concentrating solely on head reconstruction.  An example of our image preprocessing pipeline is illustrated in Fig.~\ref{fig:datapre}.

\paragraph{Train-Test Split.}
In the NeRSemble dataset, we use monocular videos from the 8-th camera as the input, only capturing the head from the front view. we evaluate head avatar reconstruction and animation quality in two settings:
1) \emph{novel view synthesis}: driving a reconstructed avatar with seen head poses and expressions during training, and rendering it from 15 hold-out viewpoints. 
2) \emph{novel expression synthesis}: driving a reconstructed avatar with unseen head poses and expressions during training, rendering it from 5 nearby hold-out views,~\ie cameras 6–10.  In ~\cref{tab:nersemble}, we provide detailed statistics about the used sequences in Nersemble and train/val/test split.

The statistics of the monocular video data are summarized in  ~\cref{tab:monocular}. Since monocular videos captured on commodity devices lack corresponding ground truths for novel view renderings, we only evaluate avatar animation performance in the quantitative comparisons, by applying pose and expression parameters from those unseen frames during training.
\begin{table}[t]
        \vspace{-2mm}
	\renewcommand\arraystretch{1.3}
        \setlength{\tabcolsep}{4pt}
	\begin{center}
        \footnotesize
		\begin{tabular}{*{4}{c}}
            \toprule
            \multirow{2}*{$\#$Sequence Names} & \multicolumn{3}{c}{$\#$timesteps $\times$ $\#$cams} \\
 			\cmidrule(lr){2-4} 
                & Train    & Novel view & Novel expression   \\
                \midrule
                \midrule
                017 EXP-5  & $208 \times 1$  & $208 \times 15$  &  $52 \times 5$ \\
                037 EXP-8  & $203 \times 1$  & $203 \times 15$ &  $50 \times 5$ \\
                055 EMO-4  & $105 \times 1$  & $105 \times 15$ &  $26 \times 5$ \\
                074 EXP-5   & $179 \times 1$  & $179 \times 15$ & $44 \times 5$ \\
                134 EMO-1   & $79 \times 1$  &  $79 \times 15$ & $19 \times 5$ \\
                165 EXP-8   & $189 \times 1$  & $189 \times 15$ & $47 \times 5$  \\
                221 EXP-5   & $147 \times 1$  & $147 \times 15$ & $36 \times 5$  \\
                251 EMO-1   & $51 \times 1$  & $51 \times 15$ &  $12 \times 5$ \\
                264 EMO-1   & $75 \times 1$  & $75 \times 15$ &  $18 \times 5$ \\
                304 EMO-1   & $127 \times 1$ &  $127 \times 15$&  $31 \times 5$ \\
                417 EMO-4   & $124 \times 1$  & $124 \times 15$ & $ 31 \times 5$ \\
                460 EXP-4   & $124 \times 1$  &  $124 \times 15$ & $30 \times 5$  \\
                \bottomrule
        \end{tabular}
        \vspace{1mm}
        \caption{\textbf{Statistics of the train/val/test splits used for NeRSemble sequences.} For each sequence, we use 80\% of timesteps for the training and validation datasets. We select the 8th camera (front-facing) for the train split, while all remaining cameras are used for novel-view evaluation (validation set). The novel-expression evaluation is conducted by selecting 5 nearby cameras for the remaining 20\% of timesteps (test set).} 
        \label{tab:nersemble}
        \end{center}
\end{table}
\begin{table}[t]
	\renewcommand\arraystretch{1.3}
        \setlength{\tabcolsep}{4pt}
	\begin{center}
        \footnotesize
		\begin{tabular}{*{4}{c}}
            \toprule
            \multirow{2}*{$\#$Sequence Names} & \multicolumn{3}{c}{$\#$timesteps} \\
 			\cmidrule(lr){2-4} 
                & Train     & Test   \\
                \midrule
                \midrule
                wojteck-1  &  760 &  2678 \\
                person0004   &  450  & 1050 \\
                subject1   & 229  & 48 \\
                subject2   & 312  & 83 \\
                subject3   & 139  & 34  \\
                subject4   & 440  & 154 \\
                \bottomrule
        \end{tabular}
        \vspace{1mm}
        \caption{\textbf{Statistics of the train/test splits used for the Monocular Video dataset.} To effectively evaluate the ability of our method to represent unseen regions of the head, we select training frames with limited head rotation. The remaining frames, which contain unseen poses and expressions, are used as the test set.} 
        \label{tab:monocular}
        \end{center}
\end{table}

\section{Implementations}
\label{SecImple}
\subsection{Monocular Head Tracking}
We track the FLAME \cite{FLAME} parameters using the VHAP-tracker~\cite{qian2024versatile} proposed in~\cite{GaussianAvatars}. Given a monocular video we optimize both shared parameters (shape, albedo map, diffuse light) and per-timestep parameters (pose, translation, expression). 
The tracking algorithm is divided into three stages: \begin{enumerate*}[label=(\roman*)]
    \item initialization stage;
    \item sequential optimization stage;
    \item global optimization stage.
\end{enumerate*}
The tracking process begins with an initialization stage, performed on the first frame of the video, which sets up all the aforementioned parameters.
Following this, a sequential optimization stage is applied to each successive frame of the video. In this stage, the parameters of each frame are optimized for 50 iterations, using the previous timestep as initialization.
Finally, the tracking parameters are refined through a global optimization stage, where a random frame is sampled at each iteration, for a total of 30 epochs.

 The tracking is performed by minimizing a combination of multiple energy terms: \begin{enumerate*}[label=(\roman*)]
    \item a photometric energy term, computed between the rendered image and the ground-truth one;
    \item a landmarks energy term, which computes the distance between the projected 2D FLAME~\cite{FLAME} landmarks and the 2D landmarks predicted by an off-the-shelf detector~\cite{Zhou_2023_star};
    \item temporal energy terms, applied on the per-timestep parameters, which ensure smoothness over time;
    \item regularization energy terms, applied on all FLAME~\cite{FLAME} parameters.
\end{enumerate*}
We revised the loss weights for the smoothness terms as: $\lambda_{smooth,transl}=\expnumber{3}{4}$, $\lambda_{smooth,rot}=\expnumber{3}{3}$, $\lambda_{smooth,jaw}=4.0$, $\lambda_{smooth,eyes}=1.0$, $\lambda_{smooth,expr}=0.5$.
For all the remaining hyper-parameters we refer to the original work~\cite{qian2024versatile}.

We use NVDiffRast~\cite{Laine2020diffrast} as the differentiable mesh renderer and the FLAME 2023 version~\cite{FLAME} with the additional 168 triangles to represent the teeth, as proposed by~\cite{GaussianAvatars}. 
\subsection{Multi-view Latent Head Diffusion}
\begin{figure*}[!htp]
    \centering
    \includegraphics[width=\linewidth]{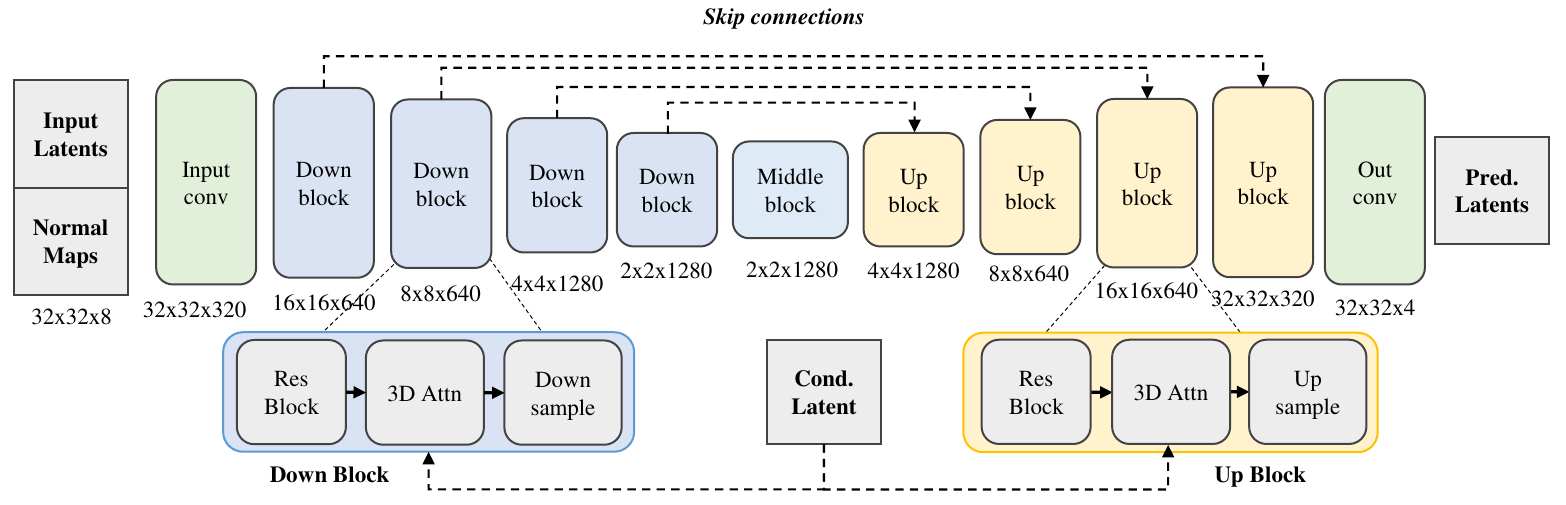}
    \caption{\textbf{Network architecture of our multi-view head latent diffusion model.} The denoiser network is built on a 2D U-Net architecture with attention blocks. The input consists of multi-view image latents concatenated with VAE latents of normal maps rendered from the FLAME mesh.     The 3D Attention block enforces 3D consistency by applying cross-attention across all views.  It also incorporates the input image latent into the denoising process, effectively preserving the identity and appearance details of the input portrait.
    }
    \label{fig:denoiser}    
\end{figure*}

In Fig.~\ref{fig:denoiser}, we show the network architecture details of our multi-view head latent diffusion. 
The denoiser network is based on a 2D U-Net~\cite{Unet} with attention blocks~\cite{Attention}.  The U-Net comprises four Down Blocks, one Middle Block, and four Up Blocks. Each Down Block contains a Residual block, a 3D Attention block, and a Downsampling layer. The Middle Block is composed of a Residual block and a 3D Attention block. The Up Block mirrors the Down blocks but with Upsampling layers.

\subsection{Gaussian Regularizations}
The position regularization term ensures that Gaussians remain close to their attached triangles during optimization through:
\begin{equation}
\vspace{-1.5mm}
    \set{L}_{pos} = \| max(\Vec{\mu}, \epsilon_{pos}) \|_2
\end{equation} where $\epsilon_{pos} = 1$ serves as the threshold, allowing small positional errors within the scaling of the attached triangle. The scale regularization term mitigates the formation of large Gaussians, which could lead to jittering problems due to small rotations of triangles. 
\begin{equation}
\vspace{-1.5mm}
    \set{L}_{scale} = \| max(\Vec{s}, \epsilon_{scale}) \|_2
\end{equation} 
It will be disabled when the local scale of the Gaussian w.r.t the attached triangle is less than $\epsilon_{pos}=0.6$.

\subsection{Ablation Studies of Gaussian Avatars Fusion}
In the main paper, we evaluate various diffusion priors for novel view constraints, demonstrating the effectiveness of our face-specific multi-view diffusion priors for 3D  Gaussian head reconstruction from monocular videos. Here, we provide additional details on the implementation of our ablation studies. We use six sequences from the NeRSemble dataset, including '055 EXP-5', '098 EMO-1', '134 EMO-1', '165 EMO-1', '221 EXP-8', and '417 EMO-4'. 

\paragraph{No diffusion.} This variant does not apply any priors to constrain novel view renderings. It is implemented using GA~\cite{GaussianAvatars} with $SH=$0.

\paragraph{Pretrained Stable Diffusion.} This variant randomly renders a novel view at each iteration and refines it using Stable Diffusion 2.1, guided by ControlNet~\cite{ControlNet} with normal maps. The iteratively denoised images serve as pseudo-ground truths.

\paragraph{Personalized Stable Diffusion.} Instead of using a pre-trained model, this variant employs DreamBooth~\cite{Dreambooth} to fine-tune the U-Net and text encoder with a learning rate of $\expnumber{5}{-6}$ for 500 iterations. The iteratively denoised images are used as pseudo-supervision.

\paragraph{Pose-conditioned multi-view diffusion.} This variant  uses pose embedding-conditioned multi-view diffusion models to generate pseudo-ground truths.

\paragraph{Raymap-conditioned multi-view diffusion.} This variant  uses ray map-conditioned multi-view diffusion models to generate pseudo-ground truths.

\paragraph{Our multi-view diffusion using Score Distillation Sampling (SDS) loss.} Instead of using iteratively denoised images as pseudo-ground truths, this variant employs SDS loss~\cite{Dreamfusion} based on single-step denoising.

\paragraph{Ours without latent upsampler $\times$2.} We remove the pretrained latent upsampler $\times$2. Then the resolution of pseudo ground truths is 256$\times$256.

\paragraph{Ours without 3D-aware denoising.}  The pseudo-ground truths generated by diffusion models rely on multi-view renderings of 3D Gaussian avatars, embedding 3D awareness into the diffusion process. To eliminate this 3D awareness, we set the time step to 0 when adding noise. Comparisons in Tab.~\ref{tab:compare_3daware} demonstrate the benefits of 3D-aware denoising.

\begin{table}[!htbp]
    \setlength{\tabcolsep}{1.3pt}
	\begin{center}
        \footnotesize
        {
		\begin{tabular}{*{7}{c}}
            \toprule
            \multirow{2}*{Method} & \multicolumn{3}{c}{Novel Views}   & \multicolumn{3}{c}{Novel Expressions}  \\
        \cmidrule(lr){2-4} \cmidrule(lr){5-7}
                 & LPIPS $\downarrow$  & PSNR $\uparrow$ & SSIM $\uparrow$   
                 & LPIPS $\downarrow$  & PSNR $\uparrow$ & SSIM $\uparrow$   \\ 
            \midrule
                Ours, w/o 3D-aware & 0.119 & 21.52& 89.71  & \bf0.079 & 25.20 & 91.99 \\
                Ours final   & \bf 0.118 & \bf 21.82 & \bf 89.87 & \bf 0.079 & \bf 25.39 & \bf 92.02 \\
                \bottomrule
        \end{tabular}
        }
        \caption{Ablation study of 3D-aware denoising.}
        \label{tab:compare_3daware}
        \end{center}
\end{table}
\section{Results of Multi-view Head Diffusion}
\label{SecMVHDRes}

\begin{figure*}[!htp]
    \centering
    \includegraphics[width=\linewidth]{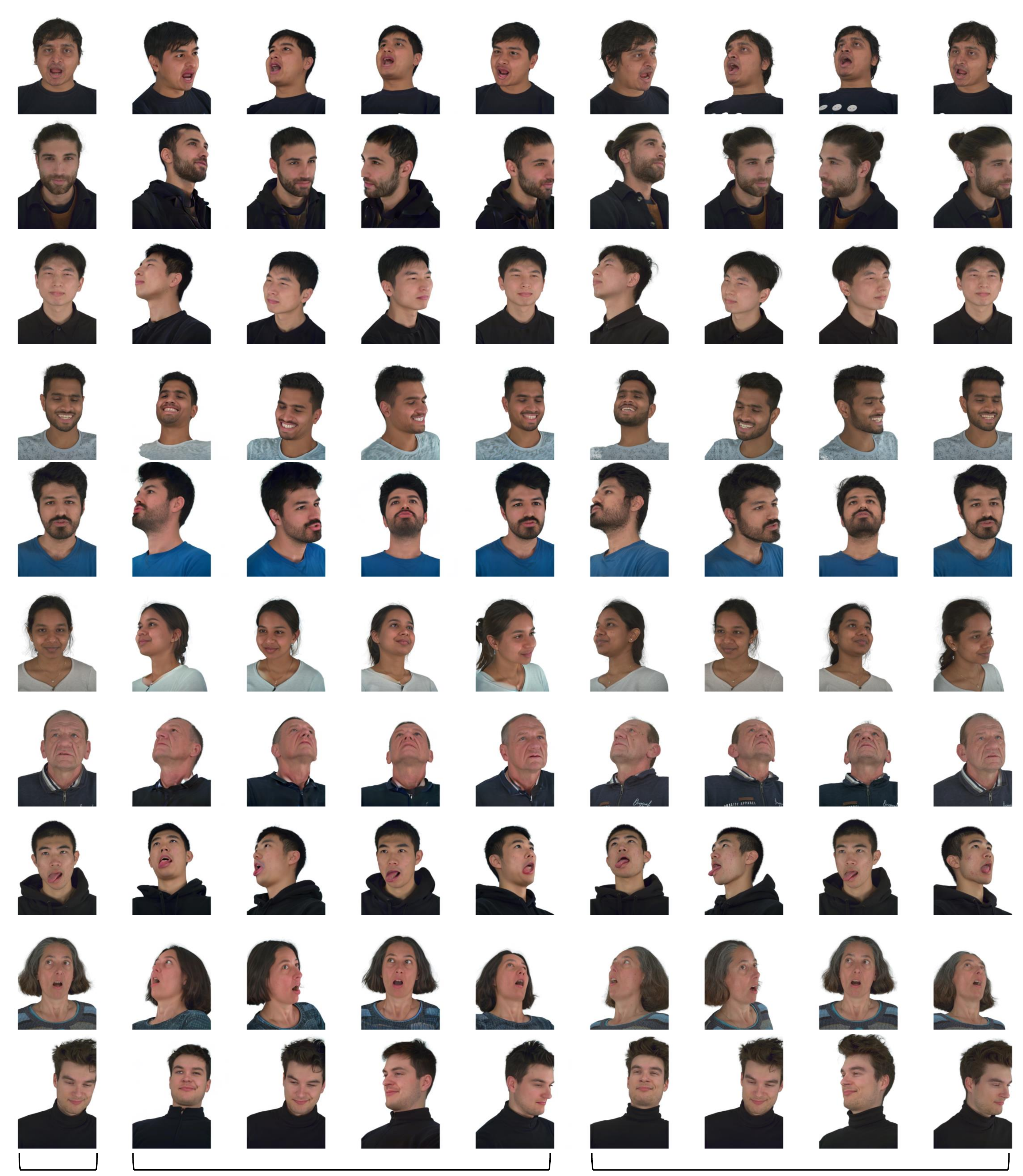}
    \begin{tabular*}{\linewidth}{@{\extracolsep{\fill}}*{9}{c}}
        Input
        & 
        & Generation Sample
        &
        & 
        & Ground Truth
        &
        &\\
    \end{tabular*}
    \caption{\textbf{Generation sample results of the multi-view head latent diffusion model.} Given a single image as input, our method can generate identity-preserved, view-consistent multi-view portrait images. }
    \label{fig:mvldm_sample}    
\end{figure*}

In Fig.~\ref{fig:mvldm_sample}, we showcase the sampling results from our multi-view head diffusion model. The model generates four view-consistent images from a single input image while effectively preserving facial identity and appearance. This demonstrates the model's capability to synthesize coherent and identity-preserving novel views.

\section{Additional Comparisons}
\label{SecAddRes}
\subsection{NeRSemble Dataset} 
\begin{figure*}[!htp]
    \centering
    \includegraphics[width=\linewidth]{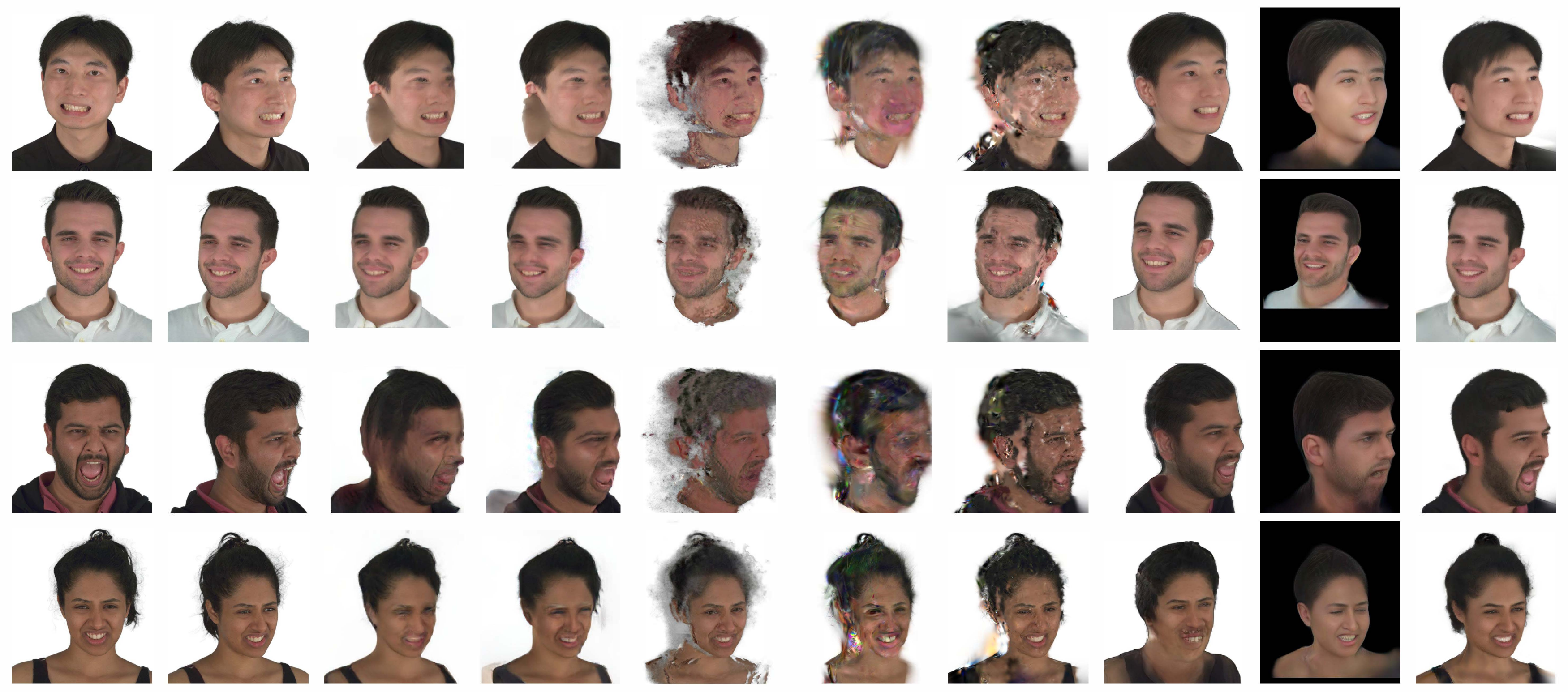}
    \begin{tabularx}{\textwidth}{XXXXXXXXXX}
        \centering \small Input & \centering \small GT   & \centering \small PanoHead & \centering \small SphereHead & \centering \small INSTA & \centering \small FlashAvatar 
        & \centering \small GA & \centering \small P4D-v2 & \centering \small GAGAvatar & \centering \small Ours
    \end{tabularx}
    \caption{\textbf{Additional results on novel view synthesis from monocular videos from the NeRSemble dataset.} 
    {Our method demonstrates robust reconstruction of less observed regions (e.g., side facial areas), maintains facial identities across viewpoints, and consistently produces more plausible and view-consistent renderings from hold-out views. }
}
    \label{fig:novelview_supple}    
\end{figure*}

In Fig.~\ref{fig:novelview_supple}, we provide additional qualitative comparisons on dynamic head avatar reconstruction from monocular videos sampled from NerSemble dataset~\cite{Nersemble}.

\subsection{Monocular Videos on Commodity Devices}
\begin{figure*}[!htbp]
    \centering
    \includegraphics[width=\linewidth]{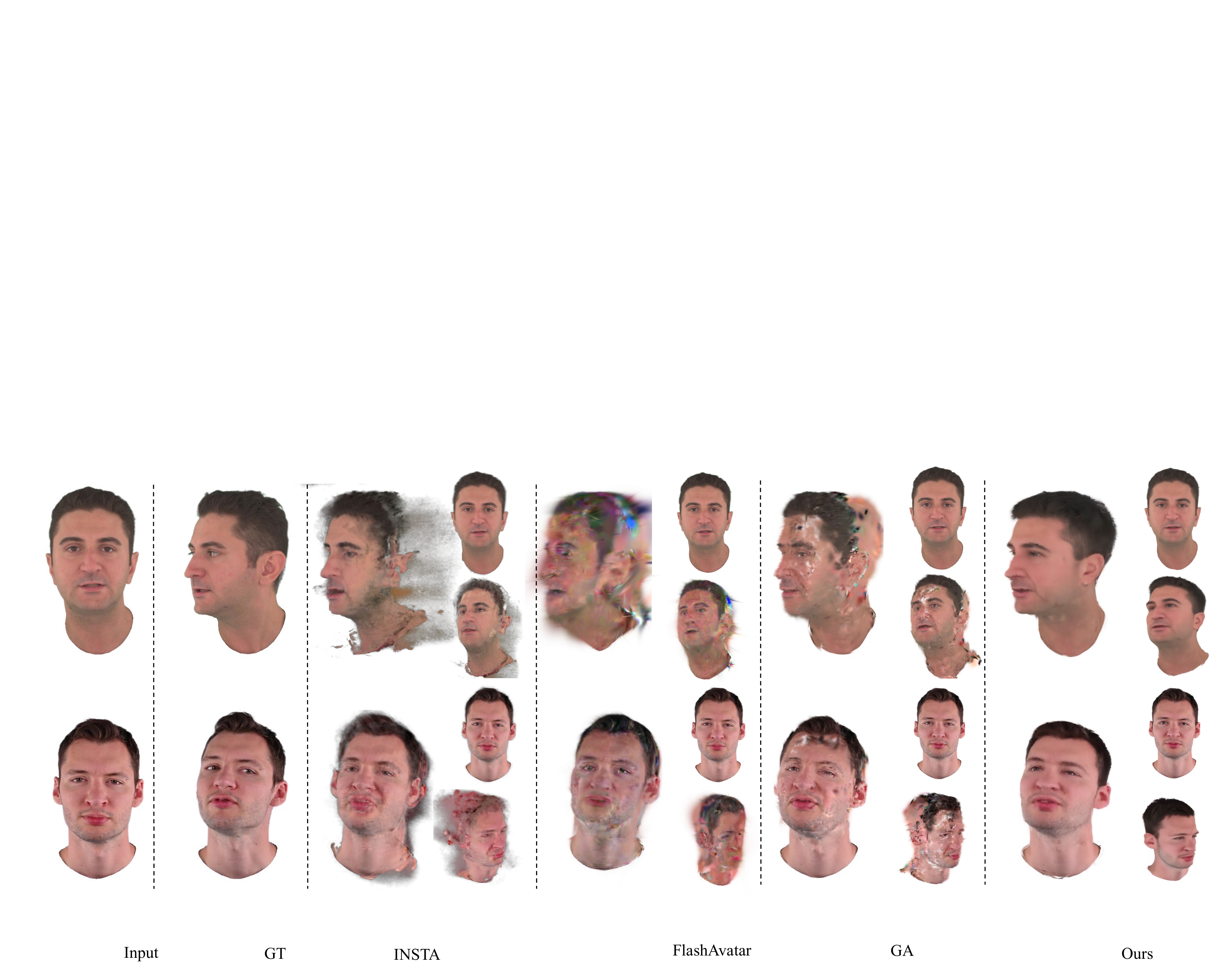}
    \begin{tabular*}{\linewidth}{@{\extracolsep{\fill}}*{9}{c}}
        (a) Input
        & (b) GT
        & (c) INSTA
        & (d) FlashAvatar
        & (e) GA
        & (f) Ours\\
    \end{tabular*}
    \caption{\textbf{Additional comparisons of head avatar reconstruction from monocular videos on commodity devices.} We present novel expression animation results using unseen frames from the monocular videos during training. Additionally, we display the fitting results for the input frame (top right) and novel view renderings of the same frame (bottom right). While all methods accurately fit observed frames in front-facing sequences with limited head poses, baseline methods fail to generalize to novel views and poses due to the absence of effective priors for less unobserved regions.  }
    \label{fig:novelview_mono_supple}    
\end{figure*}

{ In Fig.~\ref{fig:novelview_mono_supple}, we provide additional qualitative comparisons against INSTA~\cite{INSTA}, FlashAvatar~\cite{FlashAvatar}, and GA~\cite{GaussianAvatars} on monocular videos captured by commodity devices.
}

\begin{figure*}[!htp]
    \centering
    \includegraphics[width=\linewidth]{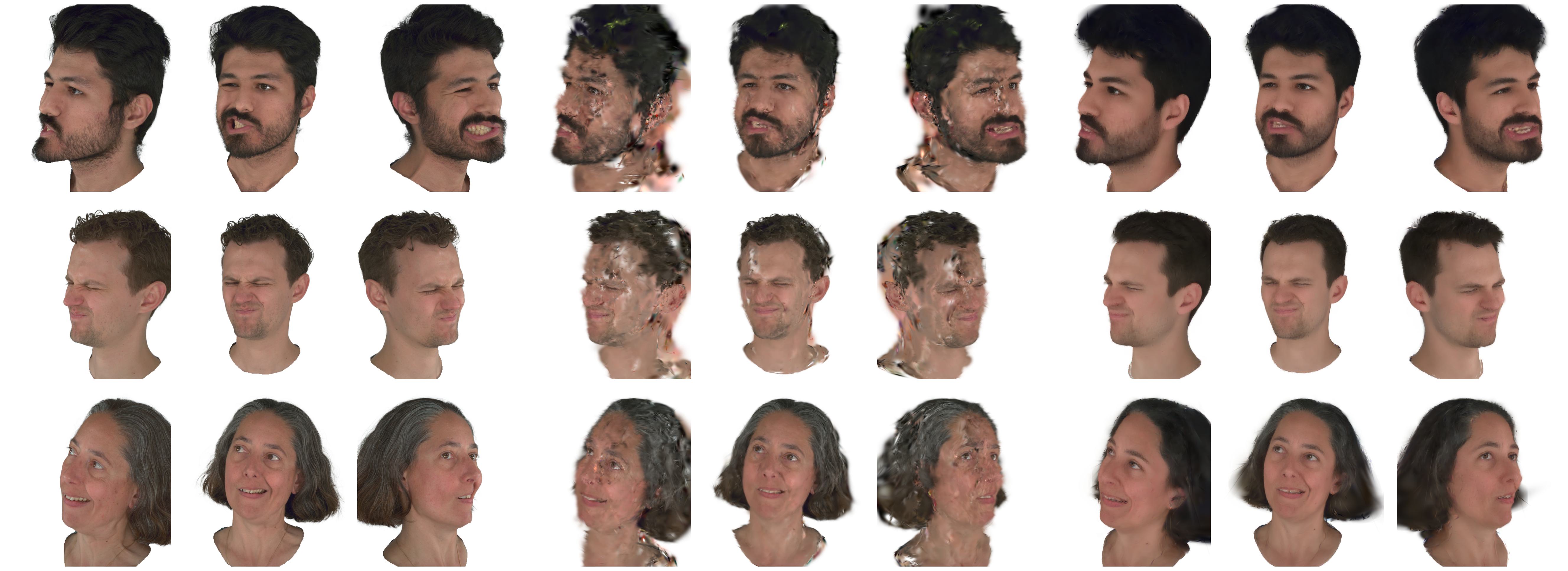}
    \begin{tabular*}
        {\linewidth}{@{\extracolsep{\fill}}p{0.32\linewidth}p{0.32\linewidth}p{0.32\linewidth}}
        \centering (a) Driving Expression (Ground truth)  & \centering (b) Gaussian Avatars & \centering (c) Ours \\
    \end{tabular*}
    \vspace{-6mm}
    \caption{\textbf{Self-reenactment from monocular videos on the NeRSemble dataset.}
    We use the tracked FLAME pose and expressions of a driving sequence to animate the reconstructed Gaussians. We show three novel view renderings for each reenactment result. Our method demonstrates more plausible head animations through more detailed face reconstruction, such as wrinkles, and faithfully produces view-consistent head renderings from different novel viewpoints.
    }
    \label{fig:selfnact}
\end{figure*}
\begin{figure*}[!htp]
    \centering
    \includegraphics[width=\linewidth]{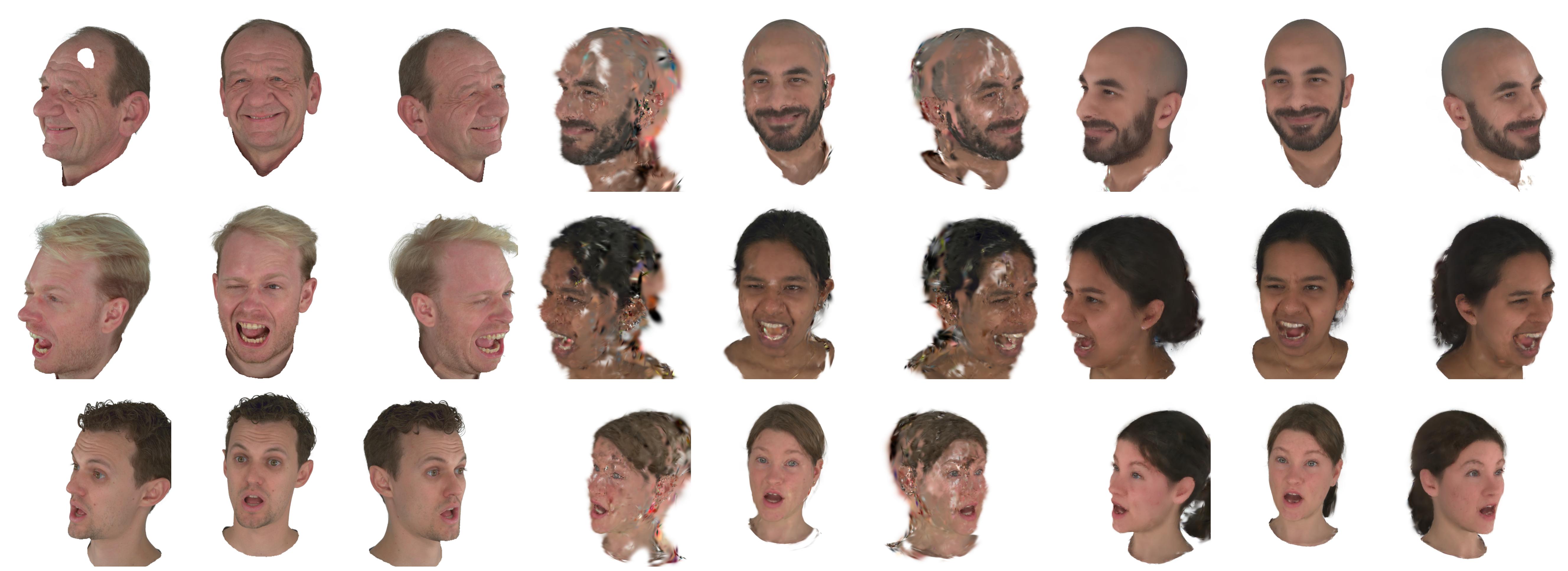}
    \begin{tabular*}
        {\linewidth}{@{\extracolsep{\fill}}p{0.32\linewidth}p{0.32\linewidth}p{0.32\linewidth}}
        \centering (a) Driving Expression (Reference)  & \centering (b) Gaussian Avatars & \centering (c) Ours \\
    \end{tabular*}
    \vspace{-6mm}
    \caption{\textbf{Cross-reenactment from monocular videos on the Nersemble dataset.} 
      We show three novel view renderings for each reenactment result. Our method outperforms Gaussian Avatars by showcasing more vivid expression transfers and more plausible renderings around the mouth.}
    \label{fig:crossnact}
\end{figure*}

\subsection{Self- \& Cross-Reenactment}
We show the self and cross reenactment results of our method and Gaussian Avatars in Fig.~\ref{fig:selfnact} and~\ref{fig:crossnact}.

\subsection{Robustness Analysis}  
To demonstrate the robustness of our method with sparse input data, we evaluate reconstruction performance across different frame numbers in the input video. We use the '104 EMO-1' sequence from the NeRSemble dataset, which contains 56 frames in the input. To reduce the frame count, we sample keyframes at uniform intervals. For instance, for an 8-frame input, we select frames at timesteps 0, 7, 14, 21, 28, 35, 42, and 55; for a single-frame input, we use only the 28th frame. 
As shown in Fig.~\ref{fig:robust}
, our method can maintain stable quantitative performance with as few as 8 frames, while GaussianAvatars drops dramatically. This highlights the resilience of our method to limited observations.

\begin{figure*}[!htp]
    \centering
        \includegraphics[width=.33\linewidth]{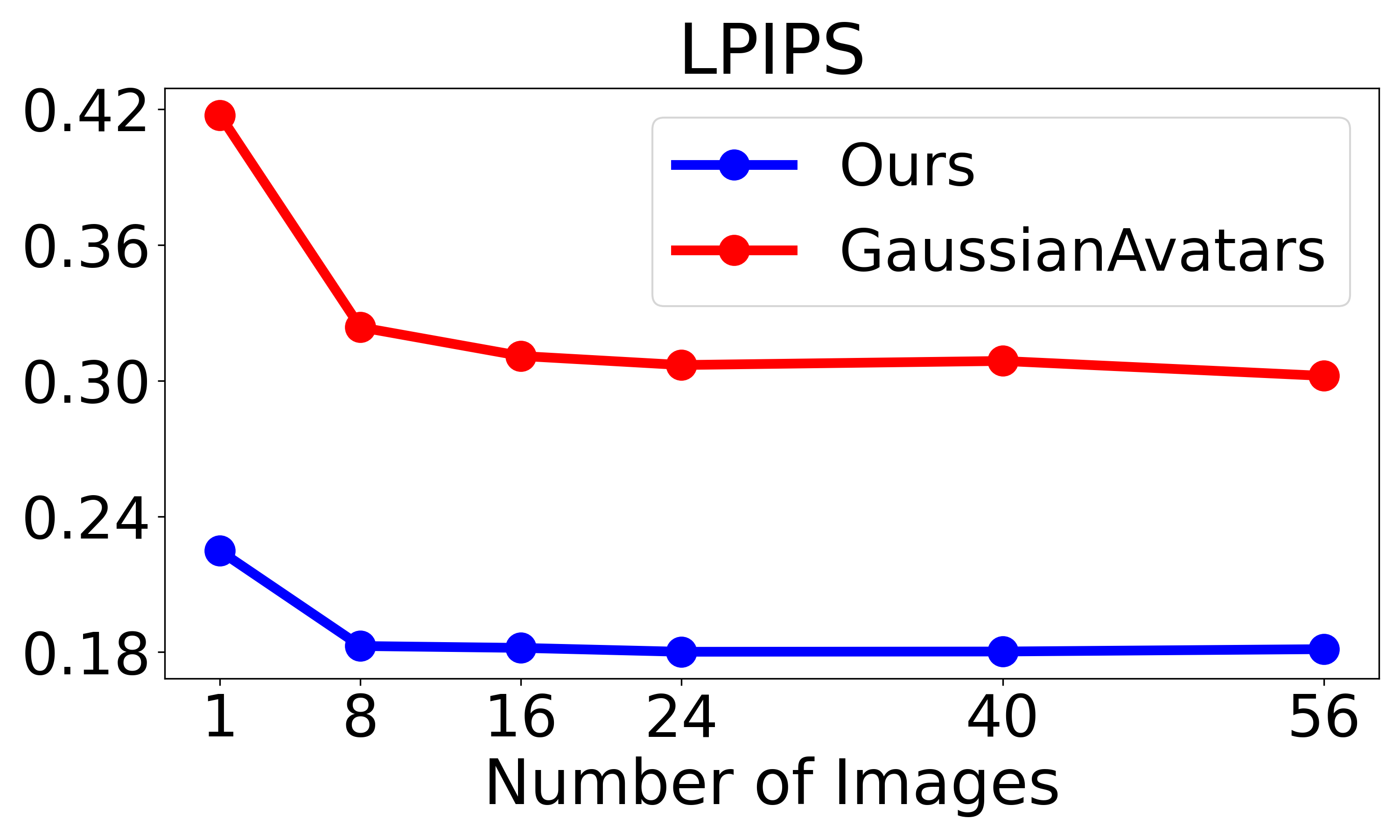}
            \hfill
        \includegraphics[width=.33\linewidth]{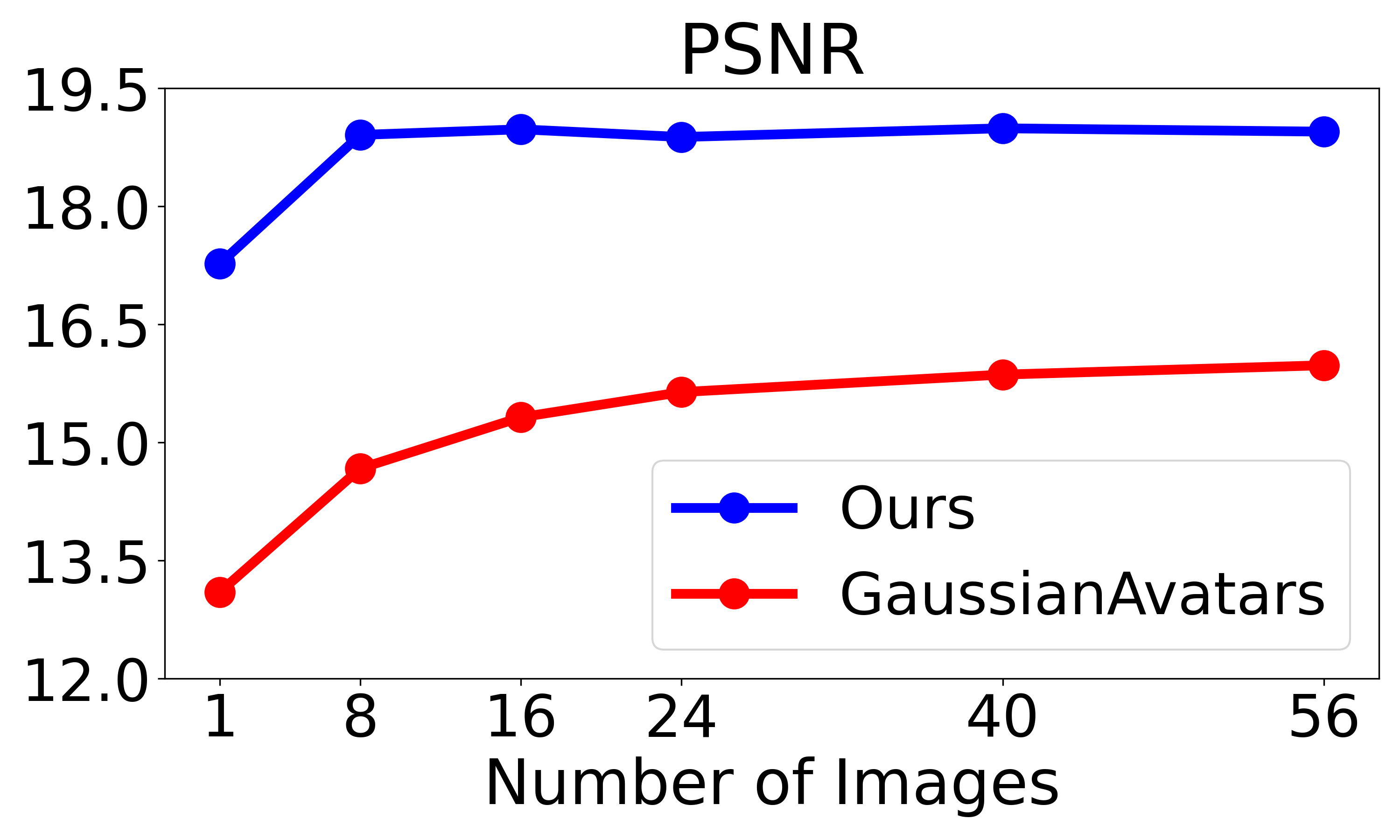}
            \hfill
        \includegraphics[width=.33\linewidth]{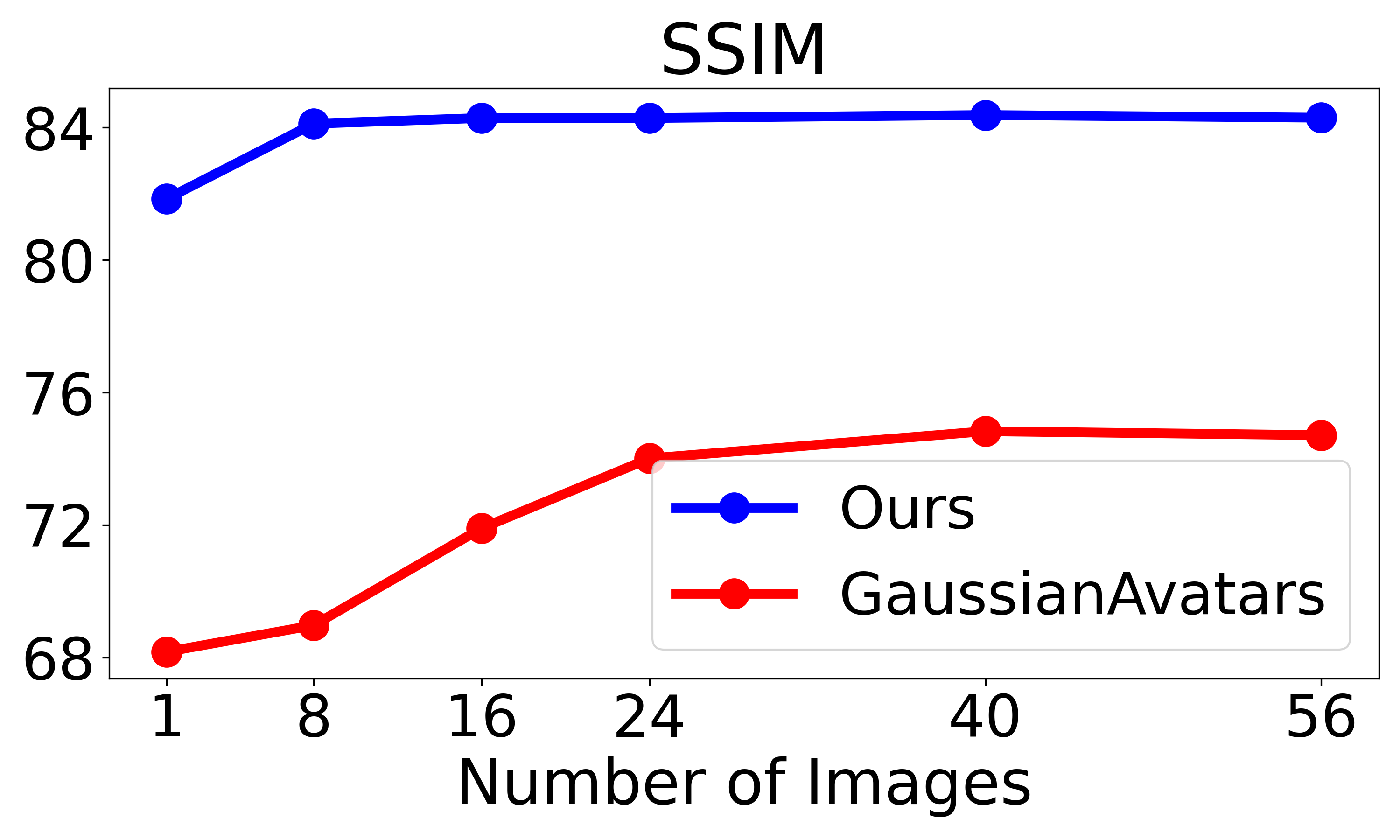}
    \caption{\textbf{Robustness analysis to the number of frames in the input monocular video.}}
    \label{fig:robust}
\end{figure*}
In Fig.~\ref{fig:robust}, we present qualitative results from a robustness analysis conducted with varying numbers of frames in the input monocular videos. Our approach consistently achieves photorealistic novel view rendering across various sequence lengths, even with only 8 frames as input.
\begin{figure*}[t]
    \centering
    \includegraphics[width=\linewidth]{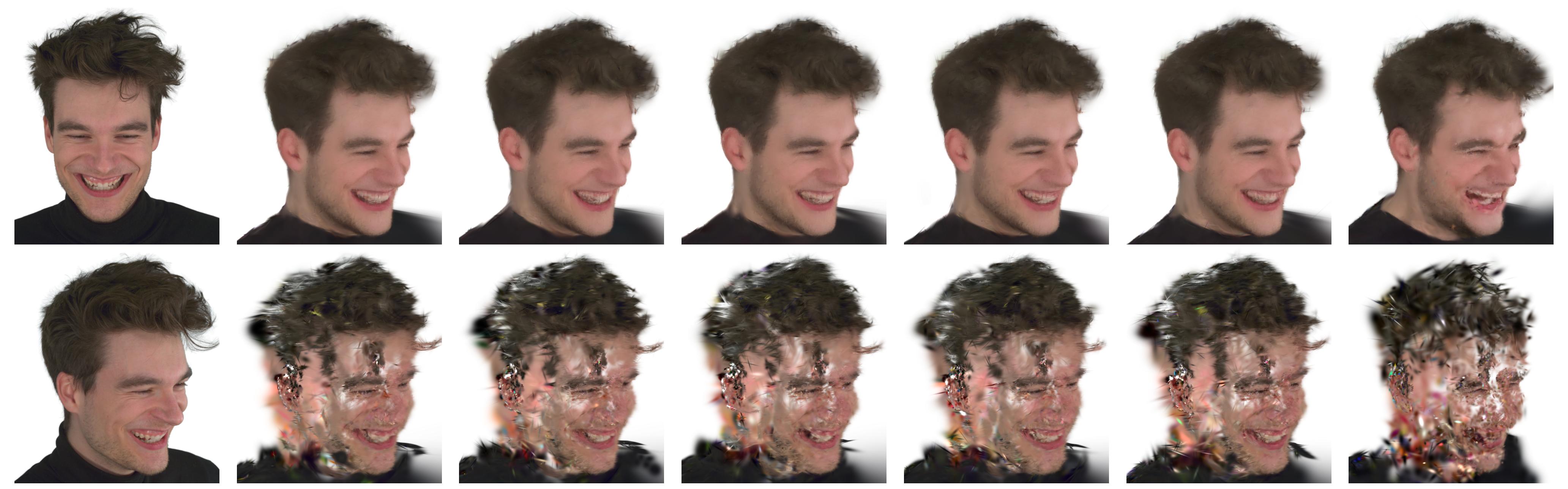}
\begin{tabular*}{\linewidth}{@{\extracolsep{\fill}}p{0.12\linewidth}p{0.12\linewidth}p{0.12\linewidth}p{0.12\linewidth}p{0.12\linewidth}p{0.12\linewidth}p{0.12\linewidth}}
\centering (a) & \centering (b) & \centering (c) & \centering (d) & \centering (e) & \centering (f) & \centering (g) \\
\end{tabular*}
    \vspace{-6mm}
    \caption{
    \textbf{Robustness analysis to the number of frames in the input monocular video.}
    (a) Input view (top) $\&$ Ground truth (bottom); (b) 56 frames; (c) 40 frames; (d) 24 frames; (e) 16 frames; (f) 8 frames; (g) 1 frame.
    Compared to GaussianAvatars~\cite{GaussianAvatars}, our method demonstrates robust reconstruction of novel view synthesis even with as few as eight frames, highlighting its robustness to limited observations.
    }
    \label{fig:robust}
    \vspace{-4mm}
\end{figure*}

\section{Ethical Discussion and Negative Impacts}
\label{SecImpact}
The creation of photorealistic and animatable head avatars from an input video poses several ethical challenges and significant risks related to the possible malevolent usage of this technology. One major concern is the potential for misuse in creating deepfakes, which are highly realistic but fake videos that can be used to spread misinformation, manipulate public opinion, or damage reputations. Additionally, this technology can lead to privacy violations, as individuals' likenesses can be replicated without their consent, leading to unauthorized use in various contexts. There is also the risk of identity theft, where malicious actors could use these avatars to impersonate others for fraudulent activities. Moreover, the psychological impact on individuals who see their digital likeness used inappropriately can be profound, causing distress and harm. Our commitment is to promote the responsible and ethical use of this technology, and we are firmly against any malicious usage that aims to harm individuals or communities.

\end{document}